\documentclass{article}

\usepackage{microtype}
\usepackage{graphicx}
\usepackage{subfigure}
\usepackage{booktabs} 
\usepackage{colortbl}
\usepackage{makecell}

\usepackage{hyperref}



\usepackage[accepted]{icml2025}

\usepackage{amsmath}
\usepackage{amssymb}
\usepackage{mathtools}
\usepackage{amsthm}
\usepackage{multirow}
\usepackage{amsmath}
\usepackage{graphicx} 
\usepackage{array} 

\usepackage[capitalize,noabbrev]{cleveref}

\theoremstyle{plain}
\newtheorem{theorem}{Theorem}[section]

\newtheorem{lemma}[theorem]{Lemma}

\theoremstyle{definition}
\newtheorem{definition}[theorem]{Definition}

\theoremstyle{remark}

\usepackage[textsize=tiny]{todonotes}
\def\ie{{\em i.e.}}
\def\eg{{\em e.g.}}

\def\sampler{\texttt{Sampler}}
\def\similar{\texttt{sim}}

\icmltitlerunning{Preserving AUC Fairness in Learning with Noisy Protected Groups}

\begin{document}

\twocolumn[
\icmltitle{Preserving AUC Fairness in Learning with Noisy Protected Groups}



\icmlsetsymbol{equal}{*}

\begin{icmlauthorlist}
\icmlauthor{Mingyang Wu}{equal,pu}
\icmlauthor{Li Lin}{equal,pu}
\icmlauthor{Wenbin Zhang}{sch}
\icmlauthor{Xin Wang}{yyy}
\icmlauthor{Zhenhuan Yang}{comp}
\icmlauthor{Shu Hu}{pu}
\end{icmlauthorlist}

\icmlaffiliation{pu}{Department of Computer and Information Technology, Purdue University, West Lafayette, USA}
\icmlaffiliation{yyy}{Department of Epidemiology and Biostatistics, University at Albany, State University of New York, New York, USA}
\icmlaffiliation{comp}{Amazon, New York, USA}
\icmlaffiliation{sch}{Knight Foundation School of Computing and Information Sciences, Florida International University, Miami, USA}

\icmlcorrespondingauthor{Shu Hu}{hu968@purdue.edu}

\icmlkeywords{Machine Learning, ICML}

\vskip 0.3in
]



\printAffiliationsAndNotice{\icmlEqualContribution} 

\begin{abstract}
The Area Under the ROC Curve (AUC) is a key metric for classification, especially under class imbalance, with growing research focus on optimizing AUC over accuracy in applications like medical image analysis and deepfake detection. 
This leads to fairness in AUC optimization becoming crucial as biases can impact protected groups. While various fairness mitigation techniques exist, fairness considerations in AUC optimization remain in their early stages, with most research focusing on improving AUC fairness under the assumption of clean protected groups. 
However, these studies often \textit{overlook} the impact of noisy protected groups, leading to fairness violations in practice. 
To address this, we propose the \textit{first} robust AUC fairness approach under noisy protected groups with fairness theoretical guarantees using distributionally robust optimization. 
Extensive experiments on tabular and image datasets show that our method outperforms state-of-the-art approaches in preserving AUC fairness. The code is in \url{https://github.com/Purdue-M2/AUC_Fairness_with_Noisy_Groups}.
\end{abstract}

\section{Introduction}
\label{submission}


The \textit{Area Under the ROC Curve} (AUC) \cite{hanley1982meaning} is one of the most widely used performance metrics in classification tasks, particularly when addressing challenges such as class imbalance or uncertain relative costs of false positives and false negatives. It provides a measure of a classifier’s ability to distinguish between classes across all possible decision thresholds, making it especially relevant in domains like information retrieval \cite{cortes2003auc}, medical image analysis \cite{yuan2021large}, and deepfake detection \cite{pu2022learning}. In particular, in deepfake detection, misclassifying fake content as real can lead to the widespread dissemination of misinformation, potentially undermining public trust, manipulating discourse, or enabling fraud. AUC is thus preferred over fixed-threshold metrics, as it captures model performance comprehensively across varying operational settings.

This has motivated numerous studies \cite{yang2021deep, kumagaiauc, guo2022robust, zhang2023doubly} focusing on training AI models to maximize AUC rather than relying on traditional loss functions (\eg, cross-entropy loss), as this approach directly optimizes a metric better aligned with the desired application outcomes. By doing so, models achieve improved performance in scenarios where distinguishing between classes with high sensitivity and specificity is essential.

The optimization of AUC in machine learning models necessitates a focus on fairness, particularly in light of growing concerns that algorithmic decisions often exacerbate inequities faced by vulnerable groups defined by sensitive attributes such as gender and race, also known as \textit{protected groups}. Recent studies \cite{caton2024fairness,kenfack2024a} have highlighted how machine learning models, if left unchecked, can perpetuate or worsen biases in allocation decisions, leading to unfavorable outcomes for these groups. To address this, a range of bias mitigation techniques \cite{hu2024fairness, lin2024preserving,tian2025fairdomain,kollias2024domain,ju2024improving} has been developed, including methods that focus on statistical fairness metrics derived from confusion matrices. 
However, fairness considerations in AUC optimization \cite{yang2023minimax, yao2023stochastic} remain an early stage of exploration despite its significance in many applications and most of them focus on the \textit{group-level} AUC fairness.

\begin{figure*}[t]
  \centering
  \includegraphics[width=1.0\linewidth]{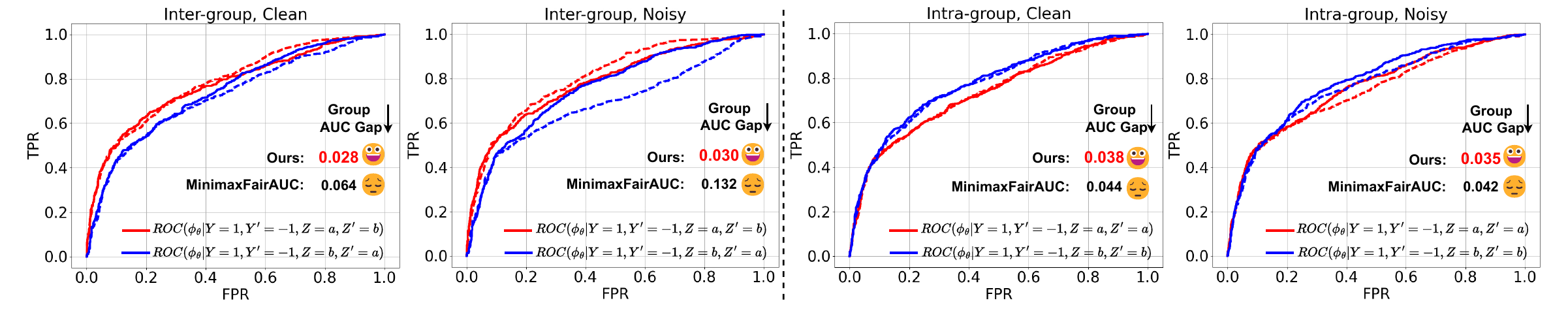}
  \vspace{-3mm}
  \caption{\it \small Illustrative inter-/intra-group AUC dicrepancy examples of existing MinimaxFairAUC method \cite{yang2023minimax} (dashed curves) and our method (solid curves) on \texttt{Default}~\cite{yeh2009comparisons} dataset with noisy levels 0 and 0.3, respectively. Notations are defined in Section \ref{sec:methodology}. In general, our method is better than MinimaxFairAUC in preserving AUC fairness, demonstrating robustness to noisy groups.}
  \label{fig:intro}
\end{figure*}

The group-level fairness for AUC leads to three categories of metrics, each addressing different aspects of disparate impacts. First, \textit{intra-group} AUC \cite{beutel2019fairness, yao2023stochastic} focuses constraining both positive and negative examples to the same group. 
Second, \textit{inter-group} AUC \cite{beutel2019fairness,kallus2019fairness, yao2023stochastic} considers ranking fairness between groups, evaluating the metric by comparing positive examples from one group to negative examples from another. 
Lastly, a few works \cite{yang2023minimax, yao2023stochastic} have sought to considering both intra-group and inter-group AUC fairness during the learning process.

Nevertheless, existing AUC fairness studies often \textbf{overlook} the reliability of protected group information. This raises a crucial question: \textit{Can AUC fairness notions be accurately measured or effectively enforced when the protected group data is noisy or unreliable?} Noisy protected group labels are prevalent in many scenarios. For instance, survey participants may intentionally obfuscate their responses due to concerns about privacy, fear of disclosure, or potential discrimination, leading to response biases \cite{krumpal2013determinants}. Similarly, in deepfake datasets, demographic annotations are often inferred using deep learning models \cite{lin2024ai}. However, the accuracy of these annotations is inherently limited, as the true demographic information cannot be verified or tracked when the faces are AI-generated. 
Our practical evaluation (see Fig. \ref{fig:intro}) reveals that training with traditional AUC fairness under noisy protected group labels can result in significant group AUC gap in model deployment. \textit{This highlights the critical need for designing a robust approach to ensure reliable AUC fairness.} 

In this work, we propose the \textbf{first} robust AUC fairness approach for learning under noisy protected groups, providing theoretical fairness guarantees. We begin by conducting experiments to illustrate the adverse effects of noisy protected group labels on existing AUC fairness methods. Next, we introduce a novel and general AUC fairness metric that accounts for both intra-group and inter-group AUC. Based on this metric, we formulate a new learning objective for AUC fairness using a distributionally robust optimization (DRO) framework \cite{duchi2021learning}, which bounds the Total Variation (TV) distance between clean and noisy group distributions. We also provide a theoretical analysis demonstrating that our approach ensures fairness even under noisy protected group labels. To estimate the TV distance bound, we reformulate it in terms of noisy label ratios and propose an empirical estimation method leveraging pre-trained multi-modal foundation models. Finally, we design an efficient stochastic gradient descent-ascent (SGDA) algorithm to optimize the proposed learning objective, enhancing both AUC fairness and the model’s generalization capabilities in deployment scenarios.
Our key contributions are:
\begin{enumerate}
    \item We present the first experimental analysis of the impact of noisy protected group labels on the existing AUC fairness learning method.
    
    \item We introduce a novel AUC fairness metric and propose the first approach to preserve AUC fairness under noisy protected groups with theoretical guarantees. 
    \item Extensive experiments on tabular and image datasets show that our method surpasses state-of-the-art approaches across applications like socioeconomic analysis and deepfake detection.  
\end{enumerate}  


\section{Related Work}

\textbf{AUC-based Fairness}. 
A pioneering effort by \citet{dixon2018measuring} introduced the Pinned AUC metric for text classification tasks, which involves resampling the data so that each of the two groups constitutes half of the dataset, followed by calculating the AUC difference on the resampled data. Building on this foundation, \citet{beutel2019fairness} proposed intra-group and inter-group AUC metrics for recommender systems, assessing whether clicked items are ranked above unclicked items both within and across protected groups. Their method also incorporated a regularization term to reduce ranking unfairness. To address disparities across groups, \citet{kallus2019fairness} developed the cross-AUC (xAUC) metric, which identifies systematic biases where positive instances from one group may be ranked below negative instances from another.

In the context of general pairwise ranking, \citet{narasimhan2020pairwise} proposed maximizing AUC under fairness constraints, further advancing the exploration of cross-group AUC fairness. Other works, such as \citet{vogel2021learning}, focused on fairness defined directly in terms of the ROC curve, employing regularization to balance overall AUC performance with group-level fairness requirements. More recently, \citet{yang2023minimax} introduced a minimax fairness framework that simultaneously addresses intra-group and inter-group AUC disparities using a Rawlsian approach, supported by an efficient optimization algorithm with proven convergence guarantees. Similarly, \citet{yao2023stochastic} proposed a scalable and efficient stochastic optimization framework for AUC-based fairness constraints, demonstrating its ability to balance classification performance and fairness in both online and offline learning scenarios. \textit{However, all of the aforementioned works assume clean and accurately labeled protected groups, disregarding the impact of noisy group labels that are common in real-world datasets. Addressing this limitation is the central focus of our paper.}

\textbf{Fairness with Noisy Protected Groups}. Group fairness methods typically assume accurate knowledge of protected group labels, but in practice, these labels are often noisy or unreliable. Enforcing fairness constraints based on such noisy labels fails to guarantee fairness with respect to the clean labels \cite{Gupta2018}. To address this issue, \citet{lahoti2020fairness} proposed an adversarial reweighting approach that leverages correlations between non-protected features, task labels, and potentially unobserved group membership, demonstrating improved fairness under label uncertainty in tabular datasets. However, extending this approach to image data poses significant challenges. 

Under the more conservative assumption of no protected group information, \citet{hashimoto2018fairness} applied distributionally robust optimization (DRO) to enforce what \citet{lahoti2020fairness} termed Rawlsian Max-Min fairness. Although DRO-based methods can achieve reasonable fairness results without explicit protected group labels, they are often less effective than approaches that incorporate such information. Building on \citet{hashimoto2018fairness}, \citet{wang2020robust} introduced a maximum total variation distance bound in the DRO procedure, offering the first fairness framework with guarantees under noisy protected-group labels. Further advancements include the work of \citet{celis2021fair}, who proposed an optimization framework with provable guarantees on both accuracy and fairness for classifiers trained with noisy protected attributes. Similarly, \citet{mehrotra2022fair} introduced a novel method for improving fair rankings in the presence of noisy group labels. Additionally, \citet{ghazimatin2022measuring} and \citet{ghosh2023fair} conducted empirical evaluations of fairness approaches under noisy sensitive information. \textit{However, the aforementioned approaches are either not generalizable or are unsuitable for direct application to pairwise ranking optimization problems. These limitations leave AUC fairness largely unexplored in these contexts. 
}


\section {Motivation}
To demonstrate the impact of noisy protected group levels on AUC fairness, we conduct experiments on the tabular Adult dataset (for socioeconomic analysis) \cite{asuncion2007uci} and the image-based FF++ dataset (for deepfake detection) \cite{rossler2019faceforensics++, lin2024preserving}. Noise is introduced by flipping a portion (ranging from 0 to 50\%) of the protected group labels (\eg, gender) in the training sets. For the Adult dataset, we evaluate the performance of the latest AUC fairness method, MinimaxFairAUC \cite{yang2023minimax}, while for the FF++ dataset, we assess the state-of-the-art fairness approach, DAW-FDD \cite{ju2024improving}. All experiments follow the original settings specified in these methods. We use AUC fairness violation as the evaluation metric, which quantifies the maximum gap between group-level (intra-group or inter-group) AUC and the overall AUC. Then, we present the mean and standard deviation scores on the test sets over three random runs for each noise setting.

\begin{figure}[t]
\small
\centering
\begin{tabular}{cc}
    \includegraphics[width=0.22\textwidth]{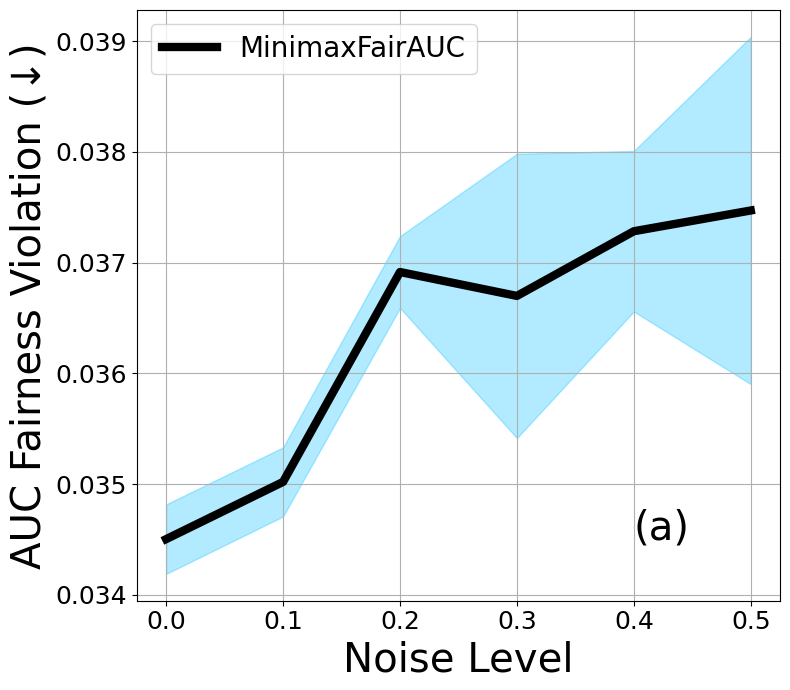} &
    \includegraphics[width=0.22\textwidth]{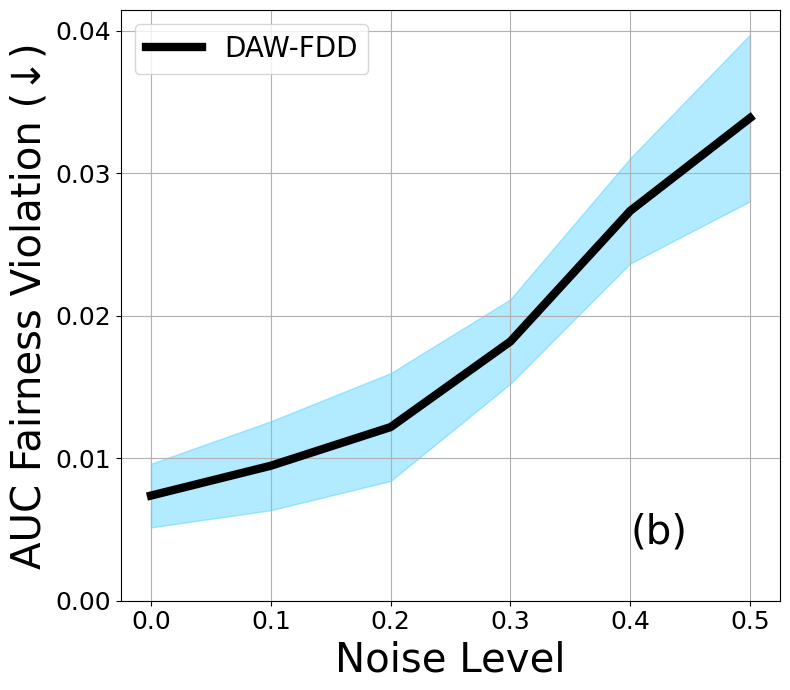} 
\end{tabular}
\vspace{-4mm}
\caption{ \it Impact of noisy protected group labels on AUC fairness violation (lower values indicate better AUC fairness) in two scenarios: (a) Socioeconomic Analysis and (b) Deepfake Detection. Mean value is shown in black line. The standard deviation is shown in blue background, where three random runs for each noise level.
}
\label{fig:motivation}
\end{figure}

As shown in Fig. \ref{fig:motivation}, AUC fairness violation increases significantly as the accuracy of protected group annotations declines across all three scenarios. Specifically, Fig. \ref{fig:motivation}(a) demonstrates that conventional AUC fairness enhancement method is highly sensitive to noisy protected group labels in the training data. Similarly, while DAW-FDD is designed to improve general fairness, Fig. \ref{fig:motivation}(b) reveals it struggles in maintaining AUC fairness under noisy group conditions. These findings underscore the critical need for robust AUC fairness approaches capable of tolerating noisy groups. 

\section{Methodology}\label{sec:methodology}

In this section, we introduce a novel robust AUC fairness approach to address the challenges discussed in the previous section. Let $X \in \mathcal{X} \subseteq \mathbb{R}^d$ represent the random variable for input features, $Y \in \mathcal{Y} = \{ \pm 1\}$ denote the binary label, and $Z \in \mathcal{Z} = \{1, \ldots, m\}$ represent the random protected group, where $m$ is the total number of protected groups. We define a scoring function $\phi_{\theta}: \mathcal{X} \to \mathbb{R}$, parameterized by $\theta \in \Theta$, which maps the input features to a real-valued score.

\subsection{AUC Fairness}
\textbf{Overall AUC}: The overall AUC quantifies the probability that a model correctly ranks a positive example (\eg, $X $) higher than a negative example (\eg, $X'$). It is formally defined as:  
\begin{equation}
\begin{aligned}
AUC(\theta) = \mathbb{E}[\mathbb{I}_{[\phi_\theta(X) > \phi_\theta(X')]} \mid Y = 1, Y' = -1],
\end{aligned}
\end{equation}  
where $\mathbb{I}_{[a]}$ is the indicator function, which equals 1 if $a$ is true and 0 otherwise. To optimize the AUC score, it is common practice to minimize the complementary AUC risk \cite{hanley1982meaning}, given by:  
\begin{equation}
\begin{aligned}
    L_{AUC}(\theta) &= 1 - AUC(\theta) \\
    &= \mathbb{E}[\mathbb{I}_{[\phi_\theta(X) \leq \phi_\theta(X')]} \mid Y = 1, Y' = -1].
\end{aligned}
\end{equation}

\textbf{Group-level AUC}. Following \cite{yang2023minimax}, the group-level AUC is defined as: 
\begin{equation}
\begin{aligned}
    &AUC_{z,z'}(\theta)\\
    &=\mathbb{E}[\mathbb{I}_{[\phi_\theta(X)>\phi_\theta(X')]}|Y=1, Y'=-1, Z=z, Z'=z'].
\end{aligned}
\end{equation}
This metric captures the AUC score for comparisons between positive examples from group $z$ and negative examples from group $z'$, reflecting a pairwise dependence with respect to the groups $Z, Z' \in \mathcal{Z}$. When $z = z'$, we refer to it as \textit{intra-group} AUC, which measures ranking performance within a single group. Conversely, when $z \neq z'$, it is termed \textit{inter-group} AUC, assessing the ranking performance across different groups.

\textbf{AUC Fairness Metric}. Then, we propose the target AUC fairness metric as follows:
\begin{equation}
\begin{aligned}    h(\theta)=&\mathbb{I}_{[\phi_\theta(X)>\phi_\theta(X')]}\mathbb{I}_{[Y=1]}\mathbb{I}_{[Y'=-1]}\\&-\mathbb{E}[\mathbb{I}_{[\phi_\theta(X)>\phi_\theta(X')]}\mathbb{I}_{[Y=1]}\mathbb{I}_{[Y'=-1]}].
\end{aligned}
\end{equation}
Building on this, we define the group-level AUC fairness functions as: 
\begin{equation}
\begin{aligned}    
g_{z,z'}(\theta)&=\mathbb{E}[h(\theta)|Z=z, Z'=z']\\
&=AUC_{z,z'}(\theta)-AUC(\theta), \ \ \forall z, z'\in \mathcal{Z}.
\end{aligned}
\label{eq:fairness-constraint}
\end{equation}
The above formulation quantifies the gap ($|g_{z,z'}(\theta)|$) between any group-level AUC score and the overall AUC score. To achieve AUC fairness, it suffices to enforce the constraint $g_{z,z'}(\theta) \leq 0$ $\forall z, z' \in \mathcal{Z}$, ensuring that all group-level AUCs are close to the overall AUC. This, in turn, reduces disparities among group-level AUCs, fostering a more equitable model performance across groups. This formulation also establishes meaningful connections with several existing fairness measures. For instance, when $z = z'$, it generalizes to intra-group pairwise fairness, as studied in \citet{beutel2019fairness, yao2023stochastic}. Conversely, when $z \neq z'$, it aligns with inter-group pairwise fairness, as explored in \citet{beutel2019fairness, kallus2019fairness, yao2023stochastic}. Furthermore, our AUC fairness function is closely related to the Rawlsian principle of justice \cite{rawls2001justice}, particularly the Rawlsian AUC fairness framework proposed by \citet{yang2023minimax}. Specifically, if we focus solely on maximizing the smallest group-level AUC among all groups, without considering the second term (\ie, the overall AUC) in Eq. (\ref{eq:fairness-constraint}).

However, solely emphasizing and enforcing the fairness constraint can lead to a trivial solution where $AUC_{z,z'}(\theta) = AUC(\theta) = 0.5$, which reflects no discriminatory power in the model. To prevent this, it is necessary to simultaneously maximize the overall AUC score while ensuring AUC fairness. It is worth noting that AUC is not only a fairness-related metric but also a key performance measure for evaluating trained models. Consequently, directly optimizing AUC can enhance model performance, as demonstrated in various domains, including medical image analysis \cite{yuan2021large} and deepfake detection \cite{pu2022learning}. Instead of maximizing the overall AUC, we minimize its corresponding AUC risk directly. This leads us to formulate the following constrained AUC fairness problem:
\begin{equation}
\begin{aligned}    
\min_\theta L_{AUC}(\theta), \  \text{s.t.} \ \ g_{z,z'}(\theta)\leq 0, \forall z, z'\in \mathcal{Z}.
\end{aligned}
\label{eq:AUC_fairness}
\end{equation}

\subsection{Robust AUC Fairness}
While minimizing Eq. (\ref{eq:AUC_fairness}) can achieve AUC fairness under clean protected groups, it does not guarantee fairness when the protected groups are noisy. To address this, we propose a robust AUC fairness learning objective by leveraging a distributionally robust optimization (DRO) approach \cite{duchi2021learning}, inspired by \citet{wang2020robust}.

\textbf{Formulation}. Specifically, let $\widehat{Z} \in \mathcal{Z}$ be the random variable representing the noisy protected group associated with $X$. The learning objective can then be reformulated as:
\begin{equation}
\begin{aligned}    
\min_\theta L_{AUC}(\theta), \  \text{s.t.} \ \ \widehat{g}_{z,z'}(\theta)\leq 0, \forall z, z'\in \mathcal{Z},
\end{aligned}
\label{eq:AUC_fairness_noisy}
\end{equation}
where $\widehat{g}_{z,z'}(\theta)=\mathbb{E}[h(\theta)|\widehat{Z}=z, \widehat{Z}'=z']$. Next, we analyze how far a model trained with noisy protected group labels using Eq. (\ref{eq:AUC_fairness_noisy}) deviates from satisfying the fairness constraints defined for clean protected groups. Let $p$ represent the distribution of pairwise data $((X,Y), (X', Y')) \sim p$, where $Y = 1$ and $Y' = -1$. Define $p_{z,z'}$ as the distribution of pairwise data conditioned on the clean groups $Z = z$ and $Z' = z'$, such that $((X,Y), (X', Y')) \mid (Z = z, Z' = z') \sim p_{z,z'}$. Similarly, let $\widehat{p}_{z,z'}$ represent the distribution of $((X,Y), (X', Y'))$ conditioned on the noisy groups $\widehat{Z} = z$ and $\widehat{Z}' = z'$, such that $((X,Y), (X', Y')) \mid (\widehat{Z} = z, \widehat{Z}' = z') \sim \widehat{p}_{z,z'}$. To quantify the difference between $p_{z,z'}$ and $\widehat{p}_{z,z'}$, we use the Total Variation (TV) distance \cite{duchi2021learning}, denoted as $ TV(p_{z,z'}, \widehat{p}_{z,z'})$. Based on this measure, we establish the following theoretical results (the proof is provided in Appendix \ref{appendix:a1}). 

\begin{theorem}\label{theorem:fairness_gap}
Suppose a model is trained using Eq. (\ref{eq:AUC_fairness_noisy}) with noisy groups and satisfies $\widehat{g}_{z,z'}(\theta) \leq 0 \ \forall z, z' \in \mathcal{Z}$. Let $\gamma_{z,z'}$ be an upper bound on the TV distance, such that $\gamma_{z,z'} \geq TV(p_{z,z'}, \widehat{p}_{z,z'}) \ \forall z, z' \in \mathcal{Z}$. Then, the fairness measure for the clean groups will be satisfied within a slack of $\gamma_{z,z'}$ for each pairwise group, ensuring $g_{z,z'}(\theta) \leq \gamma_{z,z'} \ \forall z, z' \in \mathcal{Z}$.       
\end{theorem}

\textbf{Relaxation}. While Theorem \ref{theorem:fairness_gap} provides an upper bound for AUC fairness on the clean groups, our goal is to ensure that $g_{z,z'}(\theta) \leq 0, \forall z, z' \in \mathcal{Z}$. To achieve this, inspired by \citet{wang2020robust}, we employ a DRO approach. Specifically, any feasible solution to the following constrained optimization problem is guaranteed to satisfy the fairness constraints for the clean groups:
\begin{equation}
\begin{aligned}    
&\min_\theta L_{AUC}(\theta), \\  &\text{s.t.}  \max_{\substack{\tilde{p}_{z,z'}:TV(\tilde{p}_{z,z'}, \widehat{p}_{z,z'})\leq \gamma_{z,z'}\\\tilde{p}_{z,z'}\ll p_{z,z'}}}  \tilde{g}_{z,z'}(\theta)\leq 0, \forall z, z'\in \mathcal{Z},
\end{aligned}
\label{eq:constraint_robust_AUC_fairness}
\end{equation}
where $\tilde{g}_{z,z'}(\theta)=\mathbb{E}_{((X,Y),(X',Y'))\sim \tilde{p}_{z,z'}}[h(\theta)]$ and $\tilde{p}_{z,z'}\ll p_{z,z'}$ indicates that $\tilde{p}_{z,z'}$ is absolutely continuous w.r.t. $p_{z,z'}$.

\textbf{Reformulation}. To simplify the constrained optimization problem in Eq. (\ref{eq:constraint_robust_AUC_fairness}) and demonstrate its practical application when the true distributions are unknown, we reformulate it as a minimax problem using a Lagrangian formulation. Additionally, we replace all expectations with those over the empirical distribution derived from a dataset $\mathcal{S}:=\{(X_i, Y_i, \widehat{Z}_i)\}_{i=1}^n$ of $n$ samples.
For convenience, we denote $X_i$ as $X_i^+$ or $X_i^-$ if it has a positive or negative label, respectively. Let $n^+ = |\{X_i^+ \mid i \in [n]\}|$ and $n^- = |\{X_i^- \mid i \in [n]\}|$ represent the total number of positive and negative samples, respectively, where $[n] = \{1, \ldots, n\}$. Similarly, we denote $X_i$ as $X_i^{z+}$ or $X_i^{z-}$ if it belongs to group $z$ with a positive or negative label, respectively. Let $n^{z+}$ and $n^{z-}$ represent the total number of positive and negative samples from group $z$, respectively. In practice, we let the empirical distribution $\widehat{p}_{z,z'} \in \mathbb{R}^{n^+ \times n^-}$ be a matrix where the $(i, j)$-th entry is given by: $\widehat{p}_{z,z'}^{i,j} = \frac{1}{n^{z+}n^{z'-}}$
if the $i$-th positive sample belongs to group $z$ and the $j$-th negative sample belongs to group $z'$; otherwise, it is set to 0. 

The TV distance constraint can then be reformulated to identify an empirical distribution $\tilde{p}_{z,z'} \in \mathbb{R}^{n^+ \times n^-}$ within a ball defined by:  
$\mathbb{B}_{\gamma_{z,z'}}(\widehat{p}_{z,z'}):=\{\tilde{p}_{z,z'}:    \sum_{i=1}^{n^+}\sum_{j=1}^{n^-}|\tilde{p}_{z,z'}^{i,j}-\widehat{p}_{z,z'}^{i,j}|\leq 2\gamma_{z,z'}, \sum_{i=1}^{n^+}\sum_{j=1}^{n^-}\tilde{p}_{z,z'}^{i,j}=1, \tilde{p}_{z,z'}^{i,j}\geq0 \ \forall i\in[n^+],j\in[n^-]\}$. Denote $\lambda_{z,z'}$ as the Lagrangian multiplier associated with the pair $z, z' \in \mathcal{Z}$. Then, the empirical version of Eq. (\ref{eq:constraint_robust_AUC_fairness}) can be rewritten as:  
\begin{equation}
\begin{aligned}    
&\min_\theta\max_{\substack{\lambda_{z,z'}\geq0, \tilde{p}_{z,z'}^{i,j}\geq0}}\overline{L}_{AUC}(\theta)+\sum_{z=1}^m\sum_{z'=1}^m \lambda_{z,z'}\overline{g}_{z,z'}(\theta),\\
&\text{s.t.} \ \|\tilde{p}_{z,z'}-\widehat{p}_{z,z'}\|_{1,1}\leq 2\gamma_{z,z'}, \|\tilde{p}_{z,z'}\|_{1,1}=1, \forall z,z'\in\mathcal{Z}, 
\end{aligned}
\label{eq:robust_AUC_fairness}
\end{equation}
where $\overline{L}_{AUC}(\theta)=\frac{1}{n^{+}n^-}\sum_{i=1}^{n^+}\sum_{j=1}^{n^-}\mathbb{I}_{[\phi_{\theta}(X_i^+)\leq\phi_{\theta}(X_j^-)]}$ is the empirical form of $L_{AUC}(\theta)$, 
$\overline{g}_{z,z'}(\theta)=\sum_{i=1}^{n^{+}}\sum_{j=1}^{n^{-}}\tilde{p}_{z,z'}^{i,j}\mathbb{I}_{[\phi_{\theta}(X_i^+)>\phi_{\theta}(X_j^-)]}-\frac{1}{n^{+}n^-}\sum_{i=1}^{n^+}\sum_{j=1}^{n^-}\mathbb{I}_{[\phi_{\theta}(X_i^+)>\phi_{\theta}(X_j^-)]}$ is the empirical form of $\tilde{g}_{z,z'}(\theta)$, and $\|\cdot\|_{1,1}$ is the $L_{1,1}$-norm. In practice, the non-differentiable indicator function $\mathbb{I}$ can be replaced with a (sub)-differentiable and non-increasing surrogate loss function $\ell$. For instance, in our experiments, we replace $\mathbb{I}_{[a\leq 0]}$ with the logistic loss $\log(1 + \exp(-a))$.


\subsection{Noisy Ratio Estimation}
\textbf{Theoretical Estimation}. The learning objective in Eq. (\ref{eq:robust_AUC_fairness}) relies on the upper bound $\gamma_{z,z'}$ of the TV distance between $p_{z,z'}$ and $\widehat{p}_{z,z'}$, as demonstrated in Theorem \ref{theorem:fairness_gap}. However, in practice, $p_{z,z'}$ is typically unknown. To address this, we present a theoretical result to facilitate the estimation of $\gamma_{z,z'}$, as detailed below (proof provided in Appendix \ref{appendix:proof_lemma}):
\begin{lemma}\label{lemma:estimate_ratio}

Given a positive-negative pair group ($z,z'$), suppose the prior pairwise clean group probability $Pr[(z,z')]$ unaffected by the noise, \ie, $Pr[(Z=z,Z'=z')]=Pr[(\widehat{Z}=z,\widehat{Z}'=z')]$. Then $TV(p_{z,z'},\widehat{p}_{z,z'})\leq Pr[(Z,Z')\neq(\widehat{Z},\widehat{Z}')|(\widehat{Z}=z,\widehat{Z}'=z')]$.
\end{lemma}
According to the above Lemma, the estimation of $\gamma_{z,z'}$ can be reduced to estimating the probability $Pr[(Z,Z') \neq (\widehat{Z}, \widehat{Z}') \mid (Z = z, Z' = z')]$. In robust machine learning, various methods can be employed for this estimation. For example, an auxiliary network can be trained from scratch to estimate this probability \cite{jiang2018mentornet, yu2019does}, or an auxiliary clean dataset can be utilized for estimation \cite{kallus2022assessing, wang2020robust}. 
However, training an auxiliary network introduces additional complexity to the target model, potentially reducing its generalizability across diverse scenarios. Furthermore, many existing approaches rely on a single data modality (e.g., images) for estimation, which can limit accuracy and effectiveness. Additionally, obtaining auxiliary datasets that are well-suited to the target problem can be challenging in practice.

\textbf{Empirical Estimation}. To address these challenges, we leverage the capabilities of pre-trained multi-modal foundation models \cite{li2024multimodal, gardner2024large} to construct a noisy label detector, drawing inspiration from recent works \cite{hu2023dualcoop++, wei2024vision}, without requiring additional training. While we use image data as an illustrative example, our method can be easily adapted to other data modalities, such as tabular data, which we leave as future work. Specifically, for each image, we utilize its protected group label to design a pair of label-specific prompts: a positive prompt ($\mathcal{P}$) and a negative prompt ($\mathcal{N}$). For example, $\mathcal{P}$ is designed as ``a photo of a \{group\_name\}" and $\mathcal{N}$ as ``a photo without a \{group\_name\}," where \{group\_name\} corresponds to the protected group label ($ \widehat{Z}$) of the image. These prompts are fed into the text encoder of the CLIP model \cite{radford2021learning} to generate text feature representations $T^{\mathcal{P}}$ and $T^{\mathcal{N}}$. Simultaneously, the image is processed through CLIP's visual encoder to extract its visual feature representation $\mathcal{V}$. We then compute the cosine similarity between the visual features and the text features, resulting in $\similar(\mathcal{V}, T^{\mathcal{P}})$ and $\similar(\mathcal{V}, T^{\mathcal{N}})$. Finally, we regard the group label of the test image as clean if $\similar(\mathcal{V}, T^{\mathcal{P}}) > \similar(\mathcal{V}, T^{\mathcal{N}})$, and noisy otherwise. 

We stress that CLIP is not used for relabeling, classification, or decision-making. Instead, its strong representational power is utilized to estimate the mismatch rate between protected group labels and corresponding semantic features. This design avoids additional training (e.g., adding MLP heads), thereby maintaining the theoretical guarantees of our fairness framework. Since CLIP predictions are not perfectly reliable, using them for classification would undermine the provable robustness that our method offers. Based on this estimation, we approximate $\gamma_{z,z'}$ as follows:  
\begin{equation}
\begin{aligned}    
&Pr[(Z,Z')\neq(\widehat{Z},\widehat{Z}')|(\widehat{Z}=z,\widehat{Z}'=z')]\approx\\
&\Bigg(\Big|\Big\{(i,j)\big|\substack{\widehat{Z}_i=z,\similar(\mathcal{V}_i^+, T^{+\mathcal{P}}_i)> \similar(\mathcal{V}_i^+, T^{+\mathcal{N}}_i), \\ 
\widehat{Z}_j'=z', \similar(\mathcal{V}_j^-, T^{-\mathcal{P}}_j)\leq \similar(\mathcal{V}_j^-, T^{-\mathcal{N}}_j)} \Big\}\Big|\\
+&\Big|\Big\{(i,j)\big|\substack{\widehat{Z}_i=z,\similar(\mathcal{V}_i^+, T^{+\mathcal{P}}_i)\leq \similar(\mathcal{V}_i^+, T^{+\mathcal{N}}_i), \\  
\widehat{Z}_j'=z', \similar(\mathcal{V}_j^-, T^{-\mathcal{P}}_j)> \similar(\mathcal{V}_j^-, T^{-\mathcal{N}}_j)} \Big\}\Big|\\
+&\Big|\Big\{(i,j)\big|\substack{\widehat{Z}_i=z,\similar(\mathcal{V}_i^+, T^{+\mathcal{P}}_i)\leq \similar(\mathcal{V}_i^+, T^{+\mathcal{N}}_i), \\  
\widehat{Z}_j'=z',\similar(\mathcal{V}_j^-, T^{-\mathcal{P}}_j)\leq \similar(\mathcal{V}_j^-, T^{-\mathcal{N}}_j)} \Big\}\Big|\Bigg)\\
\big/&|\{(i,j)|\widehat{Z}_i=z, \widehat{Z}_j'=z'\}|. 
\end{aligned}
\label{eq:ratio}
\end{equation}
Here, $i \in [n^+]$ and $j \in [n^-]$, with $\mathcal{V}_i^+$ and $\mathcal{V}_j^-$ as visual feature representations, and $\mathcal{T}_i^+$ and $\mathcal{T}_j^-$ as text feature representations for positive and negative samples, respectively.


\begin{algorithm}[t]
\caption{Robust AUC Fairness}
\label{alg:proj-gda}
\begin{algorithmic}[1]
\STATE {\bf Input:} A training dataset $\mathcal{S}$ of size $n$, number of iterations $T$, batch size $b$, learning rates $\eta_\theta; \eta_\lambda; \eta_p$, and $\gamma_{z,z'}$ estimated by Eq. (\ref{eq:ratio}) \\
\STATE {\bf Initialize:} $\theta^{(1)}$, $\lambda^{(1)}_{z,z'}$, $\tilde{p}^{(1)}_{z,z'}$ for all pairs $(z,z')$

\FOR{$t = 1 \ \text{to} \ T$}
  \STATE $B = \sampler (\mathcal{S}, b)$ 
    \STATE Compute \ $\epsilon^*$ \ based on Eq. (\ref{eq:sam_optimization})
    \STATE Compute perturbed parameters:  $\overline{\theta}^{(t)} = \theta^{(t)} + \epsilon^*$
    \STATE Update $\theta$: $\theta^{(t+1)}\leftarrow \theta^{(t)}-\eta_\theta \nabla_{\theta}\mathcal{L}|_{\overline{\theta}^{(t)}}$
    
  \FOR{each $(z, z') \in \{(1,1), (1,2), \ldots, (m,m)\}$}
    \STATE Update $\lambda_{z,z'}$: $\lambda_{z,z'}^{(t+1)}\leftarrow \lambda_{z,z'}^{(t)}+\eta_\lambda \overline{g}_{z,z'}(\overline{\theta}^{(t)})$
    \STATE Update $\tilde{p}_{z,z'}$:\\ $\tilde{p}_{z,z'}^{(t+1)}\leftarrow \tilde{p}_{z,z'}^{(t)}+\eta_p\lambda_{z,z'}^{(t)}\nabla_{\tilde{p}_{z,z'}} \overline{g}_{z,z'}(\overline{\theta}^{(t)})$
    
    \STATE Project $\tilde{p}_{z,z'}^{(t+1)}$ onto $\ell_{1,1}\text{-norm}$  constraints:
    $\|\tilde{p}_{z,z'}^{(t+1)}-\widehat{p}_{z,z'}\|_{1,1}\leq 2\gamma_{z,z'}$, $\|\tilde{p}_{z,z'}^{(t+1)}\|_{1,1}=1$
  \ENDFOR
\ENDFOR
\STATE \textbf{return} $\theta^{(t^*)}$, where $t^*$ is the best iterate satisfying the constraints in Eq. (\ref{eq:robust_AUC_fairness}) with the lowest objective.
\end{algorithmic}
\end{algorithm}

\subsection{Optimization}
Finally, we develop a stochastic gradient descent-ascent (SGDA) method to solve the minimax optimization problem in Eq. (\ref{eq:robust_AUC_fairness}). To avoid the model becoming stuck in sharp and narrow minima during training, we incorporate the sharpness-aware minimization (SAM) technique \cite{foret2020sharpness} to flatten the loss landscape. This flattening is achieved by finding the optimal perturbation $\epsilon^*$ to the model parameters $\theta$, which maximizes the loss. The process is:
\begin{equation}
\begin{aligned} 
\epsilon^* &= \arg\max_{\|\epsilon\|_2 \le \nu}
      \underbrace{\bigl(\overline{L}_{AUC}+\sum_{z=1}^m\sum_{z'=1}^m \lambda_{z,z'}\overline{g}_{z,z'}\bigr)}_{\displaystyle \mathcal{L}}
      \,(\theta + \epsilon)
     \\ &\approx\quad\arg\max_{\|\epsilon\|_2 \le \nu} 
      \epsilon^\top \nabla_{\theta}\mathcal{L}\;=\;\nu\, \frac{\nabla_{\theta}\mathcal{L}}{||\nabla_{\theta}\mathcal{L}||_2}, 
\end{aligned}
\label{eq:sam_optimization}
\end{equation}
Here, $\nu$ controls the perturbation magnitude, and the approximation is derived using a first-order Taylor expansion, assuming $\epsilon$ is small. The final equation is obtained by solving a dual norm problem, where $\text{sign}$ represents the sign function, and $\nabla_\theta \mathcal{L}$ is the gradient of $\mathcal{L}$ with respect to $\theta$. Consequently, the model weights are updated by solving the following optimization problem: $\min_\theta \mathcal{L}(\theta+\epsilon^*)$.
The intuition is that perturbing the model parameters in the direction of the gradient norm maximizes the loss value, which in turn encourages the model to explore a flatter loss landscape, enhancing its generalizability. 

To ensure each mini-batch contains both positive and negative samples from all possible groups, we follow \citet{yang2023minimax} and design a sampling operator (\ie, \sampler($\cdot$, $\cdot$)) that randomly selects a mini-batch of size $b$ by stratifying the training set $\mathcal{S}$ based on the label and group attribute.  
The details of the sampling operator are outlined in Appendix \ref{append:sampling}.


The overall optimization procedure is as follows: we first initialize the model parameters $\theta$, $\lambda_{z,z'}$, and $\tilde{p}_{z,z'}$, and use the approach developed in the previous section with Eq. (\ref{eq:ratio}) to estimate $\gamma_{z,z'}$ for all possible positive-negative group pairs ($z, z'$). Next, we randomly select a mini-batch $B$ using our \sampler~ operator and perform the following steps for each iteration on $B$ (For more details, refer to Algorithm \ref{alg:proj-gda}):
\begin{itemize}
    \item Compute $\epsilon^*$ based on Eq. (\ref{eq:sam_optimization}).
    \item Update $\theta$ based on the gradient descent: 
    $\theta\leftarrow\theta-\eta_\theta \nabla_{\theta}\mathcal{L}|_{\theta+\epsilon^*}$, where $\eta_\theta$ is the learning rate.
    \item For each $(z,z')\in\{(1,1), (1,2), ..., (m,m)\}$, use gradient ascent to update $\lambda_{z,z'}$: $\lambda_{z,z'}\leftarrow \lambda_{z,z'}+\eta_\lambda \overline{g}_{z,z'}(\theta+\epsilon^*)$ and $\tilde{p}_{z,z'}$: $\tilde{p}_{z,z'}\leftarrow \tilde{p}_{z,z'}+\eta_p\lambda_{z,z'}\nabla_{\tilde{p}_{z,z'}} \overline{g}_{z,z'}(\theta+\epsilon^*)$, where $\eta_{\lambda}$ and $\eta_p$ are learning rates. Next, we project $\tilde{p}_{z,z'}$ onto $\ell_{1,1}\text{-norm}$  constraints:
    $\|\tilde{p}_{z,z'}-\widehat{p}_{z,z'}\|_{1,1}\leq 2\gamma_{z,z'}$, $\|\tilde{p}_{z,z'}\|_{1,1}=1$.  
\end{itemize}
The project can be done efficiently with \citet{duchi2008efficient}.


\section{Experiments}

\begin{table*}[t]
    \centering
    \scalebox{0.6}{
    \begin{tabular}{c|c|ccc|ccc|ccc}
    \hline
    \multirow{2}{*}{\makecell{Noise \\ Level}} 
    & \multirow{2}{*}{Method} 
    & \multicolumn{3}{c|}{Adult} 
    & \multicolumn{3}{c|}{Bank} 
    & \multicolumn{3}{c}{Default} \\ \cline{3-11} 
     &  & \multicolumn{1}{c|}{AUC$\uparrow$} 
         & Violation$\downarrow$ 
         & Min/Max$\uparrow$ 
       & \multicolumn{1}{c|}{AUC$\uparrow$} 
         & Violation$\downarrow$ 
         & Min/Max$\uparrow$ 
       & \multicolumn{1}{c|}{AUC$\uparrow$} 
         & Violation$\downarrow$ 
         & Min/Max$\uparrow$ \\ 
    \hline
    
\multirow{4}{*}{0.1}

& AUCMax 
& \multicolumn{1}{c|}{\textbf{0.9159$\pm$0.0001}}
& 0.0787$\pm$0.0018 
& 0.9115$\pm$0.0012
& \multicolumn{1}{c|}{\textbf{0.9288$\pm$0.0003}}
& 0.1565$\pm$0.0016
& 0.8559$\pm$0.0014
& \multicolumn{1}{c|}{\textbf{0.7762$\pm$0.0008}}
& 0.0701$\pm$0.0009
& 0.9076$\pm$0.0011 \\

& InterFairAUC 
& \multicolumn{1}{c|}{0.9031$\pm$0.0002}
& 0.0380$\pm$0.0013
& 0.9479$\pm$0.0007
& \multicolumn{1}{c|}{0.9068$\pm$0.0005}
& 0.0975$\pm$0.0025
& 0.9048$\pm$0.0022
& \multicolumn{1}{c|}{0.7399$\pm$0.0031}
& 0.0242$\pm$0.0003 
& 0.9649$\pm$0.0009 \\

& MinimaxFairAUC 
& \multicolumn{1}{c|}{0.9059$\pm$0.0002}
& 0.0350$\pm$0.0003
& 0.9513$\pm$0.0006
& \multicolumn{1}{c|}{0.9148$\pm$0.0006}
& 0.1246$\pm$0.0035
& 0.8795$\pm$0.0034
& \multicolumn{1}{c|}{0.7536$\pm$0.0037}
& 0.0302$\pm$0.0080
& 0.9520$\pm$0.0098 \\

& \cellcolor[HTML]{EFEFEF}Ours
& \multicolumn{1}{c|}{\cellcolor[HTML]{EFEFEF}0.9060$\pm$0.0002}
& \cellcolor[HTML]{EFEFEF}\textbf{0.0332$\pm$0.0007}
& \cellcolor[HTML]{EFEFEF}\textbf{0.9536$\pm$0.0005}
& \multicolumn{1}{c|}{\cellcolor[HTML]{EFEFEF}0.9039$\pm$0.0007}
& \cellcolor[HTML]{EFEFEF}\textbf{0.0876$\pm$0.0033}
& \cellcolor[HTML]{EFEFEF}\textbf{0.9143$\pm$0.0029}
& \multicolumn{1}{c|}{\cellcolor[HTML]{EFEFEF}0.7645$\pm$0.0017}
& \cellcolor[HTML]{EFEFEF}\textbf{0.0187$\pm$0.0012}
& \cellcolor[HTML]{EFEFEF}\textbf{0.9691$\pm$0.0019} \\

\hline
\multirow{4}{*}{0.2}

& AUCMax 
& \multicolumn{1}{c|}{\textbf{0.9112$\pm$0.0002}}
& 0.0885$\pm$0.0011 
& 0.9051$\pm$0.0008
& \multicolumn{1}{c|}{\textbf{0.9264$\pm$0.0003}}
& 0.1592$\pm$0.0026
& 0.8543$\pm$0.0022
& \multicolumn{1}{c|}{\textbf{0.7890$\pm$0.0005}}
& 0.0565$\pm$0.0011
& 0.9254$\pm$0.0013 \\

& InterFairAUC 
& \multicolumn{1}{c|}{0.9035$\pm$0.0004}
& 0.0377$\pm$0.0003
& 0.9477$\pm$0.0005
& \multicolumn{1}{c|}{0.9037$\pm$0.0003}
& 0.1071$\pm$0.0028
& 0.8960$\pm$0.0024
& \multicolumn{1}{c|}{0.7527$\pm$0.0037}
& 0.0256$\pm$0.0026
& 0.9560$\pm$0.0062 \\

& MinimaxFairAUC 
& \multicolumn{1}{c|}{0.9045$\pm$0.0003}
& 0.0369$\pm$0.0003
& 0.9489$\pm$0.0005
& \multicolumn{1}{c|}{0.9142$\pm$0.0006}
& 0.1226$\pm$0.0046
& 0.8814$\pm$0.0044
& \multicolumn{1}{c|}{0.7526$\pm$0.0039}
& 0.0267$\pm$0.0057
& 0.9558$\pm$0.0093 \\

& \cellcolor[HTML]{EFEFEF}Ours 
& \multicolumn{1}{c|}{\cellcolor[HTML]{EFEFEF}0.9039$\pm$0.0004}
& \cellcolor[HTML]{EFEFEF}\textbf{0.0328$\pm$0.0008}
& \cellcolor[HTML]{EFEFEF}\textbf{0.9539$\pm$0.0011}
& \multicolumn{1}{c|}{\cellcolor[HTML]{EFEFEF}0.9163$\pm$0.0005}
& \cellcolor[HTML]{EFEFEF}\textbf{0.1053$\pm$0.0026}
& \cellcolor[HTML]{EFEFEF}\textbf{0.8988$\pm$0.0024}
& \multicolumn{1}{c|}{\cellcolor[HTML]{EFEFEF}0.7624$\pm$0.0016}
& \cellcolor[HTML]{EFEFEF}\textbf{0.0206$\pm$0.0014}
& \cellcolor[HTML]{EFEFEF}\textbf{0.9660$\pm$0.0022} \\

\hline
\multirow{4}{*}{0.3}

& AUCMax 
& \multicolumn{1}{c|}{\textbf{0.9150$\pm$0.0001}}
& 0.0800$\pm$0.0005
& 0.9117$\pm$0.0004
& \multicolumn{1}{c|}{\textbf{0.9293$\pm$0.0005}}
& 0.1565$\pm$0.0026
& 0.8559$\pm$0.0023
& \multicolumn{1}{c|}{\textbf{0.7768$\pm$0.0005}}
& 0.0679$\pm$0.0013
& 0.9100$\pm$0.0015 \\

& InterFairAUC 
& \multicolumn{1}{c|}{0.8990$\pm$0.0008}
& 0.0379$\pm$0.0012
& 0.9482$\pm$0.0032
& \multicolumn{1}{c|}{0.9078$\pm$0.0005}
& 0.0994$\pm$0.0020
& 0.9031$\pm$0.0018
& \multicolumn{1}{c|}{0.7405$\pm$0.0027}
& 0.0217$\pm$0.0009
& 0.9684$\pm$0.0019 \\

& MinimaxFairAUC 
& \multicolumn{1}{c|}{0.8977$\pm$0.0006}
& 0.0367$\pm$0.0013
& 0.9490$\pm$0.0025
& \multicolumn{1}{c|}{0.9142$\pm$0.0006}
& 0.1207$\pm$0.0050
& 0.8833$\pm$0.0049
& \multicolumn{1}{c|}{0.7533$\pm$0.0036}
& 0.0290$\pm$0.0077
& 0.9526$\pm$0.0097 \\

& \cellcolor[HTML]{EFEFEF}Ours
& \multicolumn{1}{c|}{\cellcolor[HTML]{EFEFEF}0.9059$\pm$0.0005}
& \cellcolor[HTML]{EFEFEF}\textbf{0.0356$\pm$0.0006}
& \cellcolor[HTML]{EFEFEF}\textbf{0.9513$\pm$0.0007}
& \multicolumn{1}{c|}{\cellcolor[HTML]{EFEFEF}0.9093$\pm$0.0004}
& \cellcolor[HTML]{EFEFEF}\textbf{0.0949$\pm$0.0026}
& \cellcolor[HTML]{EFEFEF}\textbf{0.9080$\pm$0.0024}
& \multicolumn{1}{c|}{\cellcolor[HTML]{EFEFEF}0.7687$\pm$0.0011}
& \cellcolor[HTML]{EFEFEF}\textbf{0.0129$\pm$0.0011}
& \cellcolor[HTML]{EFEFEF}\textbf{0.9785$\pm$0.0014} \\

\hline
\multirow{4}{*}{0.4}

& AUCMax 
& \multicolumn{1}{c|}{\textbf{0.9131$\pm$0.0002}}
& 0.0783$\pm$0.0007
& 0.9123$\pm$0.0006
& \multicolumn{1}{c|}{\textbf{0.9275$\pm$0.0005}}
& 0.1655$\pm$0.0025
& 0.8465$\pm$0.0023
& \multicolumn{1}{c|}{\textbf{0.7795$\pm$0.0005}}
& 0.0717$\pm$0.0007
& 0.9057$\pm$0.0008 \\

& InterFairAUC 
& \multicolumn{1}{c|}{0.8987$\pm$0.0008}
& 0.0367$\pm$0.0022
& 0.9488$\pm$0.0042
& \multicolumn{1}{c|}{0.9060$\pm$0.0005}
& 0.0944$\pm$0.0030
& 0.9078$\pm$0.0029
& \multicolumn{1}{c|}{0.7524$\pm$0.0035}
& 0.0256$\pm$0.0039
& 0.9559$\pm$0.0068 \\

& MinimaxFairAUC 
& \multicolumn{1}{c|}{0.9056$\pm$0.0002}
& 0.0373$\pm$0.0007
& 0.9492$\pm$0.0003
& \multicolumn{1}{c|}{0.9146$\pm$0.0005}
& 0.1246$\pm$0.0036
& 0.8795$\pm$0.0035
& \multicolumn{1}{c|}{0.7538$\pm$0.0036}
& 0.0301$\pm$0.0080
& 0.9520$\pm$0.0098 \\

& \cellcolor[HTML]{EFEFEF}Ours
& \multicolumn{1}{c|}{\cellcolor[HTML]{EFEFEF}0.9008$\pm$0.0002}
& \cellcolor[HTML]{EFEFEF}\textbf{0.0341$\pm$0.0005}
& \cellcolor[HTML]{EFEFEF}\textbf{0.9530$\pm$0.0004}
& \multicolumn{1}{c|}{\cellcolor[HTML]{EFEFEF}0.9054$\pm$0.0004}
& \cellcolor[HTML]{EFEFEF}\textbf{0.0937$\pm$0.0034}
& \cellcolor[HTML]{EFEFEF}\textbf{0.9086$\pm$0.0028}
& \multicolumn{1}{c|}{\cellcolor[HTML]{EFEFEF}0.7552$\pm$0.0024}
& \cellcolor[HTML]{EFEFEF}\textbf{0.0246$\pm$0.0036}
& \cellcolor[HTML]{EFEFEF}\textbf{0.9580$\pm$0.0066} \\

\hline
\multirow{4}{*}{0.5}

& AUCMax 
& \multicolumn{1}{c|}{\textbf{0.9127$\pm$0.0003}}
& 0.0766$\pm$0.0009
& 0.9141$\pm$0.0008
& \multicolumn{1}{c|}{\textbf{0.9290$\pm$0.0003}}
& 0.1572$\pm$0.0017
& 0.8553$\pm$0.0014
& \multicolumn{1}{c|}{\textbf{0.7767$\pm$0.0007}}
& 0.0715$\pm$0.0015
& 0.9061$\pm$0.0017 \\

& InterFairAUC 
& \multicolumn{1}{c|}{0.9012$\pm$0.0012}
& 0.0374$\pm$0.0013
& 0.9482$\pm$0.0026
& \multicolumn{1}{c|}{0.9067$\pm$0.0005}
& 0.0947$\pm$0.0017
& 0.9076$\pm$0.0016
& \multicolumn{1}{c|}{0.7488$\pm$0.0037}
& 0.0278$\pm$0.0026
& 0.9562$\pm$0.0047 \\

& MinimaxFairAUC 
& \multicolumn{1}{c|}{0.9012$\pm$0.0008}
& 0.0375$\pm$0.0016
& 0.9482$\pm$0.0030
& \multicolumn{1}{c|}{0.9145$\pm$0.0006}
& 0.1229$\pm$0.0042
& 0.8811$\pm$0.0046
& \multicolumn{1}{c|}{0.7526$\pm$0.0037}
& 0.0271$\pm$0.0069
& 0.9553$\pm$0.0098 \\

& \cellcolor[HTML]{EFEFEF}Ours
& \multicolumn{1}{c|}{\cellcolor[HTML]{EFEFEF}0.8984$\pm$0.0006}
& \cellcolor[HTML]{EFEFEF}\textbf{0.0316$\pm$0.0002}
& \cellcolor[HTML]{EFEFEF}\textbf{0.9554$\pm$0.0004}
& \multicolumn{1}{c|}{\cellcolor[HTML]{EFEFEF}0.9044$\pm$0.0004}
& \cellcolor[HTML]{EFEFEF}\textbf{0.0876$\pm$0.0020}
& \cellcolor[HTML]{EFEFEF}\textbf{0.9145$\pm$0.0015}
& \multicolumn{1}{c|}{\cellcolor[HTML]{EFEFEF}0.7447$\pm$0.0047}
& \cellcolor[HTML]{EFEFEF}\textbf{0.0243$\pm$0.0014}
& \cellcolor[HTML]{EFEFEF}\textbf{0.9571$\pm$0.0010} \\

\hline
\end{tabular}
}
\caption{ 
\small \textit{Performance comparison across different noise levels (0.1--0.5). The numbers are reported as `Mean $\pm$ Standard Deviation.’ \textuparrow~means higher is better and \textdownarrow~means lower is better.  The best results are shown in \textbf{Bold}.}}
\label{tab:tabular}
\end{table*}

\subsection{Settings}
\textbf{Datasets.} In experiments, we train models with different methods on both tabular and image datasets. For tabular data, we conduct socioeconomic analysis on three widely used datasets in fair machine learning research \cite{donini2018empirical}: Adult (protected attribute: gender), Bank (protected attribute: age), and Default (protected attribute: gender). Each dataset is randomly split into training, validation, and test sets in a 60\%/20\%/20\% ratio. For image data, we focus on the deepfake detection task using datasets from \citet{lin2024preserving}. Specifically, we train models on the FF++ \cite{rossler2019faceforensics++} training set (protected attribute: gender) and evaluate them on the test sets of FF++, DFDC \cite{deepfakedetection2021}, DFD \cite{googledeepfakes2019}, and Celeb-DF \cite{li2020celebdf}. Further details are provided in Appendix \ref{appendix:dataset}.

\textbf{Evaluation Metrics.}
For utility, we use overall AUC as the primary model performance metric. For fairness, we measure AUC fairness violation (`Violation', lower is better), defined as the maximum absolute difference between any group-level AUC score the overall AUC score. Additionally, following \citet{yang2023minimax}, we use the Min/Max fairness metric (higher is better), defined as the ratio of the minimum to the maximum group-level AUC score. Their formulations can be found in Appendix \ref{appendix:metric}.

\textbf{Baselines.}
For socioeconomic analysis, we compare our method against AUCMax (without fairness constraint) \cite{yang2023minimax}, InterFairAUC \cite{vogel2021learning}, and  MinimaxFairAUC \cite{yang2023minimax}. For deepfake detection, we compare our method with the latest fairness methods, including DAG-FDD, DAW-FDD \cite{ju2024improving}, and PG-FDD \cite{lin2024preserving}. The comparison also includes `Original' (a backbone with cross-entropy loss). More details can be found in Appendix \ref{appendix:baselines}. 

\textbf{Implementation Details.}
All experiments are implemented in PyTorch and trained on an NVIDIA RTX A6000. For training, we set the batch size to 10,000 for socioeconomic analysis and 32 for deepfake detection, with 1,000 and 100 training epochs, respectively. We use the SGD optimizer. For socioeconomic analysis, we use 3-layer multilayer perceptron (MLP) as the model. $ \gamma $ is selected from $\{0.1, 0.2, 0.3, 0.4, 0.5\}$. For deepfake detection, we use noisy group labels, which are common in datasets like FF++ where demographic attributes are inferred. Evaluating fairness under label noise is practical and follows prior work \cite{celis2021fair, mehrotra2022fair}. We use Xception \cite{chollet2017xception} and EfficientNet-B4 \cite{tan2019efficientnet} as the detector backbones. $ \gamma = 0.02 $ is estimated using Eq. (\ref{eq:ratio}). See Appendix \ref{appendix:additional_implementation} for details.

\begin{table*}[t]
    \centering
    \scalebox{0.64}{
\begin{tabular}{c|c|ccc|ccc|ccc|ccc}
\hline
                                  &                                          & \multicolumn{3}{c|}{FF++}                                                                                                                        & \multicolumn{3}{c|}{DFDC}                                                                                                                        & \multicolumn{3}{c|}{DFD}                                                                                                                         & \multicolumn{3}{c}{Celeb-DF}                                                                                                            \\ \cline{3-14} 
\multirow{-2}{*}{Backbone}        & \multirow{-2}{*}{Method}                 & \multicolumn{1}{c|}{AUC\textuparrow}                                     & Violation\textdownarrow                               & Min/Max\textuparrow                                 & \multicolumn{1}{c|}{AUC\textuparrow}                                     & Violation\textdownarrow                               & Min/Max\textuparrow                                 & \multicolumn{1}{c|}{AUC\textuparrow}                                     & Violation\textdownarrow                               & Min/Max\textuparrow                                 & \multicolumn{1}{c|}{AUC\textuparrow}                            & Violation\textdownarrow                               & Min/Max\textuparrow                                 \\ \hline
                                  & Original                                 & \multicolumn{1}{c|}{0.9384}                                  & 0.0303                                  & 0.9652                                  & \multicolumn{1}{c|}{0.5953}                                  & 0.0319                                  & 0.9492                                  & \multicolumn{1}{c|}{0.7574}                                  & 0.0326                                  & 0.9529                                  & \multicolumn{1}{c|}{0.6660}                          & 0.1479                                  & 0.8101                                  \\
                                  & DAG-FDD                                  & \multicolumn{1}{c|}{0.9628}                                  & 0.0147                                  & 0.9776                                  & \multicolumn{1}{c|}{0.6058}                                  & 0.0229                                  & 0.9608                                  & \multicolumn{1}{c|}{0.7770}                                  & 0.0216                                  & 0.9633                                  & \multicolumn{1}{c|}{0.7059}                         & 0.1730                                  & 0.7906                                  \\
                                  & DAW-FDD                                  & \multicolumn{1}{c|}{0.9650}                                  & 0.0288                                  & 0.9702                                  & \multicolumn{1}{c|}{0.6037}                                  & 0.0201                                  & 0.9684                                  & \multicolumn{1}{c|}{0.7825}                                  & 0.0312                                  & 0.9479                                  & \multicolumn{1}{c|}{0.7092}                         & 0.1818                                  & 0.7792                                  \\
                                  & PG-FDD                                   & \multicolumn{1}{c|}{\textbf{0.9708}}                         & 0.0111                                  & 0.9714                                  & \multicolumn{1}{c|}{\textbf{0.6207}}                         & 0.0184                                  & 0.9594                                  & \multicolumn{1}{c|}{\textbf{0.8025}}                         & 0.0113                                  & 0.9846                                  & \multicolumn{1}{c|}{\textbf{0.7214}}                & 0.1412                                  & 0.8350                                  \\
\multirow{-5}{*}{Xception}        & \cellcolor[HTML]{EFEFEF}Ours    & \multicolumn{1}{c|}{\cellcolor[HTML]{EFEFEF}0.9644}          & \cellcolor[HTML]{EFEFEF}\textbf{0.0090} & \cellcolor[HTML]{EFEFEF}\textbf{0.9857} & \multicolumn{1}{c|}{\cellcolor[HTML]{EFEFEF}0.6086}          & \cellcolor[HTML]{EFEFEF}\textbf{0.0048} & \cellcolor[HTML]{EFEFEF}\textbf{0.9930} & \multicolumn{1}{c|}{\cellcolor[HTML]{EFEFEF}0.7847}          & \cellcolor[HTML]{EFEFEF}\textbf{0.0069} & \cellcolor[HTML]{EFEFEF}\textbf{0.9881} & \multicolumn{1}{c|}{\cellcolor[HTML]{EFEFEF}0.7108} & \cellcolor[HTML]{EFEFEF}\textbf{0.0729} & \cellcolor[HTML]{EFEFEF}\textbf{0.9117} \\ \hline
                                  & Original                                 & \multicolumn{1}{c|}{0.9332}                                  & 0.0209                                  & 0.9684                                  & \multicolumn{1}{c|}{0.5982}                                  & 0.0306                                  & 0.9320                                  & \multicolumn{1}{c|}{0.7593}                                  & 0.0445                                  & 0.9382                                  & \multicolumn{1}{c|}{0.6692}                         & 0.2453                                  & 0.6962                                  \\
                                  & DAG-FDD                                  & \multicolumn{1}{c|}{0.9563}                                  & 0.0100                                  & 0.9869                                  & \multicolumn{1}{c|}{0.6030}                                  & 0.0372                                  & 0.9210                                  & \multicolumn{1}{c|}{0.7706}                                  & 0.0216                                  & 0.9626                                  & \multicolumn{1}{c|}{0.7102}                         & 0.2196                                  & 0.7415                                  \\
                                  & DAW-FDD                                  & \multicolumn{1}{c|}{0.9694}                                  & 0.0169                                  & 0.9764                                  & \multicolumn{1}{c|}{0.5941}                                  & 0.0254                                  & 0.9449                                  & \multicolumn{1}{c|}{0.7756}                                  & 0.0380                                  & 0.9440                                  & \multicolumn{1}{c|}{0.7345}                         & 0.2873                                  & 0.6792                                  \\
                                  & PG-FDD                                   & \multicolumn{1}{c|}{0.9721}                                  & 0.0144                                  & 0.9784                                  & \multicolumn{1}{c|}{0.6043}                                  & 0.0235                                  & 0.9476                                  & \multicolumn{1}{c|}{0.8033}                                  & 0.0260                                  & 0.9650                                  & \multicolumn{1}{c|}{\textbf{0.7366}}                & 0.1356                                  & 0.8373                                  \\
\multirow{-5}{*}{\makecell{EfficientNet \\ -B4}} & \cellcolor[HTML]{EFEFEF}Ours     & \multicolumn{1}{c|}{\cellcolor[HTML]{EFEFEF}\textbf{0.9766}} & \cellcolor[HTML]{EFEFEF}\textbf{0.0061} & \cellcolor[HTML]{EFEFEF}\textbf{0.9907} & \multicolumn{1}{c|}{\cellcolor[HTML]{EFEFEF}\textbf{0.6172}} & \cellcolor[HTML]{EFEFEF}\textbf{0.0136} & \cellcolor[HTML]{EFEFEF}\textbf{0.9771} & \multicolumn{1}{c|}{\cellcolor[HTML]{EFEFEF}\textbf{0.8184}} & \cellcolor[HTML]{EFEFEF}\textbf{0.0135} & \cellcolor[HTML]{EFEFEF}\textbf{0.9760} & \multicolumn{1}{c|}{\cellcolor[HTML]{EFEFEF}0.7351} & \cellcolor[HTML]{EFEFEF}\textbf{0.0928} & \cellcolor[HTML]{EFEFEF}\textbf{0.8876} \\ \hline
\end{tabular}
}
\caption{\small \textit{Performance comparison on deepfake detection task. \textuparrow~means higher is better and \textdownarrow~means lower is better.}}
\label{tab:deepfake}
\end{table*}

\begin{figure*}[t]
  \centering
  \includegraphics[width=1.0\linewidth]{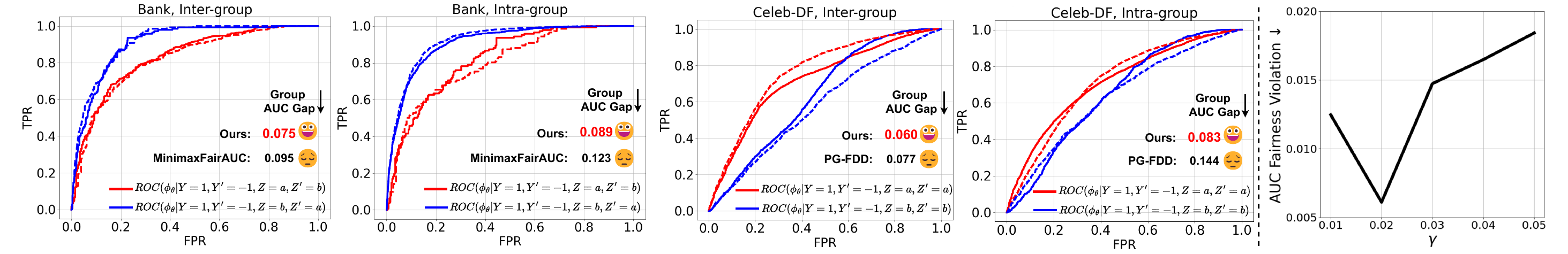}
  \vspace{-8mm}
  \caption{\it \small (\textbf{Left}) Comparison of AUC gap on inter-group and intra-group across different datasets. For the tabular dataset, we compare our method with MinimaxFairAUC on the Bank dataset under a noise level of 0.1. For the image dataset, we compare ours with PG-FDD on Celeb-DF. (\textbf{Right}) AUC fairness violation across different $\gamma$ values. $\gamma$ has been set manually from $\{0.01, 0.02, 0.03, 0.04, 0.05\}$. }
  \vspace{-2mm}
  \label{fig:roc_ablation}
\end{figure*}
\subsection{Results}
\textbf{Performance on Tabluar Data.}
For each of the three tabular datasets, we introduce noise into the protected group labels by randomly selecting a fraction $ \gamma$ of data points and flipping their labels to another group. We evaluate the performance of each method under different noise levels, conducting three random runs per dataset and reporting the mean and standard deviation in Table \ref{tab:tabular}. It is clear that our approach consistently achieves the lowest AUC fairness violation and the highest Min/Max AUC score across all datasets and noise levels, demonstrating its effectiveness in preserving fairness under noisy protected group labels. For instance, at a noise level of 0.5 on the Adult dataset, our method achieves a fairness violation of 0.0316, significantly lower than AUCMax (0.0766), InterFairAUC (0.0374), and MinimaxFairAUC (0.0375). While AUCMax attains a higher overall AUC due to the absence of fairness constraints, its fairness violation remains substantially higher. Similarly, at a noise level of 0.1 on the Default dataset, our method achieves 0.0187 fairness violation, reducing it by 5.14\% compared to AUCMax.  
Moreover, as the noise level increases, baseline methods exhibit worsening fairness violations, whereas our approach maintains superior fairness performance, underscoring its robustness in handling noisy protected group labels.

\textbf{Performance on Image Data.}
Deepfake detection datasets are inherently noisy due to inaccuracies in demographic annotations, as the protected groups of generated faces cannot be verified in practice. Thus, our proposed noisy estimation method is well-suited for estimating the noise ratio, which we determine to be 0.02 in our experiments. We fix this value for subsequent experiments and report the results in Table~\ref{tab:deepfake}.  
From the table, we find our method achieves the lowest group AUC fairness violation across all datasets, demonstrating superior fairness preservation under noisy protected groups. For instance, on FF++ (Xception backbone), our method achieves a fairness violation of 0.0090, significantly lower than PG-FDD (0.0111), DAW-FDD (0.0288), and DAG-FDD (0.0147). Similarly, in the DFDC cross-domain scenario, our method attains a violation of 0.0048, reducing fairness violation by 1.36\% compared to PG-FDD. This trend persists across DFD and Celeb-DF, where our method consistently maintains the lowest fairness violation.
While PG-FDD achieves higher AUC as it is the state-of-the-art method for fairness generalization in deepfake detection, its fairness violation remains higher than ours across all datasets. 
Overall, our results demonstrate superior fairness preservation under noisy groups across different datasets and model backbones, indicating the robustness of our method in real-world image analysis applications.

\subsection{Sensitivity Analysis}
\textbf{Inter-/Intra- Group Performance Gap.}
We examine the inter-group and intra-group AUC gaps across multiple datasets to assess the capability of our method in preserving fairness under noisy groups. As shown in the left of Fig.~\ref{fig:roc_ablation}, our method reduces the intra-group AUC gap by 3.40\% compared to MinimaxFairAUC on the Bank dataset and by 1.70\% compared to PG-FDD on the Celeb-DF dataset. 
These results demonstrate that our method effectively mitigates both intra-group and inter-group AUC disparities simultaneously in the presence of noisy labels, as evidenced by its performance across both tabular and image datasets.

\textbf{Correctness of Noise Ratio Estimation.}
To evaluate the accuracy of noise ratio estimation in the image-data scenario, we experiment with multiple noise levels ($\gamma \in \{0.01, 0.02, 0.03, 0.04, 0.05\}$), as the exact noise ratio is unknown. We apply our method to train deepfake detectors using an EfficientNet-B4 backbone and test on the FF++ dataset. As shown in Fig. \ref{fig:roc_ablation} (Right), our method achieves the lowest fairness violation at $\gamma = 0.02$, which aligns with the estimated $\gamma$ derived from Eq. (\ref{eq:ratio}). This consistency validates the correctness of our noise ratio estimation. 

\subsection{Ablation Study}
\textbf{With vs. Without SAM.} We conduct experiments on image datasets using the EfficientNet-B4 backbone without applying SAM during training and report the results in Table \ref{tab:ablation-sam}. The results show a performance decline (\eg, 0.9656 Min/Max on DFDC) compared to our full method (\eg, 0.9771), yet it still outperforms the baseline approaches in Table \ref{tab:deepfake}. This highlights the importance of SAM in enhancing AUC fairness generalization in our approach.  

\textbf{Robust vs. Non-robust.} Under the same experimental setting, we compare our full method with a version that excludes robustness by using Eq. (\ref{eq:AUC_fairness}). As shown in Table \ref{tab:deepfake}, the absence of robustness leads to a significant performance drop. For instance, on the DFDC dataset, the fairness violation increases by approximately 0.32\%, highlighting the effectiveness of our proposed robust approach. More experiments are in Appendix \ref{appendix:more_non_robust_experiments}.  

\begin{table}[t]
    \centering
    \scalebox{0.55}{
    \begin{tabular}{c|cc|ccc|ccc}
    \hline
    \multirow{2}{*}{Backbone} 
    & \multicolumn{2}{c|}{Method} 
    & \multicolumn{3}{c|}{DFDC}
    & \multicolumn{3}{c}{DFD}
     \\ \cline{2-9} 
     & Robust & SAM
     & \multicolumn{1}{c|}{AUC$\uparrow$} 
         & Violation$\downarrow$ 
         & Min/Max$\uparrow$ 
      & \multicolumn{1}{c|}{AUC$\uparrow$} 
         & Violation$\downarrow$ 
         & Min/Max$\uparrow$ 
        \\ 
    \hline

\multirow{3}{*}{\makecell{EfficientNet \\ -B4}}
& \checkmark & 
& \multicolumn{1}{c|}{0.6099}
& 0.0155
& 0.9656
& \multicolumn{1}{c|}{0.7965}
& 0.0148
& 0.9721
 \\

&  & \checkmark
& \multicolumn{1}{c|}{0.6090}
& 0.0168
& 0.9636
& \multicolumn{1}{c|}{0.8058}
& 0.0289
& 0.9611
 \\

& \cellcolor[HTML]{EFEFEF}\checkmark & \cellcolor[HTML]{EFEFEF} \checkmark 
& \multicolumn{1}{c|}{\cellcolor[HTML]{EFEFEF}\textbf{0.6172}}
& \cellcolor[HTML]{EFEFEF}\textbf{0.0136}
& \cellcolor[HTML]{EFEFEF}\textbf{0.9771}
& \multicolumn{1}{c|}{\cellcolor[HTML]{EFEFEF}\textbf{0.8184}}
& \cellcolor[HTML]{EFEFEF}\textbf{0.0135}
& \cellcolor[HTML]{EFEFEF}\textbf{0.9760}
\\

\hline
\end{tabular}
}
\caption{ 
\small \textit{Performance comparison between with and without SAM on our method. \textuparrow~means higher is better and \textdownarrow~means lower is better.  The best results are shown in \textbf{Bold}.}}
\vspace{-3mm}
\label{tab:ablation-sam}
\end{table}

\section{Conclusion}
Existing methods for enhancing AUC fairness perform well with clean protected groups but fail under noisy groups. To address this, we propose a novel DRO-based approach with theoretical fairness guarantees, achieved by bounding the TV distance between clean and noisy group distributions. We estimate this bound through theoretical analysis and then develop an empirical method leveraging pre-trained multi-modal foundation models. Finally, we design an efficient SGDA algorithm to optimize the proposed learning objective, improving both AUC fairness and model generalization. Experimental results in diverse datasets highlight the superior AUC fairness maintenance capabilities of our method in three application scenarios. 

\textbf{Limitation}. Although our approach guarantees AUC fairness, ensuring both utility and fairness simultaneously when training with noisy groups remains an open challenge. Additionally, while we use CLIP as an example of a multi-modal foundation model for noise estimation in image data, we do not imply that CLIP is directly applicable to tabular data. Rather, our statement refers to the potential to leverage domain-specific foundation models, such as those for tabular data, for analogous noise estimation tasks. Since we do not instantiate this component for tabular data in our current experiments, we acknowledge the lack of a concrete noise estimation strategy for tabular modalities as a limitation of this work and leave its exploration to future research.

\textbf{Future work}. In addition to addressing the aforementioned limitation, we aim to extend our approach to other pairwise ranking metrics \cite{yang2022algorithmic} (\eg, partial AUC, average precision) to enhance their group-level fairness.

\section*{Acknowledgements}

We thank anonymous reviewers for constructive comments. This work is supported by the U.S. National Science Foundation (NSF) under grant IIS-2434967 and the National Artificial Intelligence Research Resource (NAIRR) Pilot and TACC Lonestar6. The views, opinions and/or fndings expressed are those of the author and should not be interpreted as representing the offcial views or policies of NSF and NAIRR Pilot.

\section*{Impact Statement}

Our work addresses a critical and underexplored challenge in machine learning fairness: how to ensure AUC-based fairness when protected group labels are noisy or unreliable, a condition common in real-world datasets such as survey data and demographically annotated synthetic media. By proposing the first theoretically grounded, DRO framework tailored to AUC fairness under noisy protected groups, our research expands the practical reliability and applicability of fairness-aware machine learning.

The potential societal benefits are substantial. In domains like healthcare, finance, and digital media, where both performance and fairness are paramount, our method offers a robust solution that can help prevent algorithmic discrimination, especially under imperfect data labeling conditions. For example, it can reduce harm in risk-sensitive applications such as medical image analysis or deepfake detection by ensuring more equitable outcomes across groups.

Ethically, our approach prioritizes fairness even when data quality is compromised, an important step toward responsible AI. However, the broader implications of deploying fairness-aware systems should still be evaluated in context, especially when group identities are inferred or estimated.

We believe our contribution helps move the field toward more trustworthy and fair machine learning models, even under real-world constraints.

\nocite{langley00}

\bibliography{references}
\bibliographystyle{icml2025}

\newpage
\appendix
\onecolumn

\section{Proofs}
\subsection{Proof of Theorem \ref{theorem:fairness_gap}} \label{appendix:a1}
Our proof is inspired by \citet{wang2020robust}; however, the key distinction lies in the focus. While \citet{wang2020robust} addresses non-ranking-based fairness measures, we derive results specifically for pairwise ranking-based AUC fairness measures. First, we define TV distance between $p_{z,z'}$ and $\widehat{p}_{z,z'}$ as follows,
\begin{definition}
Let $m(x, y) = \mathbb{I}_{x \neq y}$ be a metric, and let $\pi$ represent a coupling between the probability distributions $p_{z,z'}$ and $\widehat{p}_{z,z'}$. The TV distance is defined as $TV(p_{z,z'}, \widehat{p}_{z,z'}) = \inf_{\pi} \mathbb{E}_{X, Y \sim \pi}[m(X, Y)] \ \ \text{s.t.} \int \pi(x, y) dy = p_{z,z'}(x), \int \pi(x, y) dx = \widehat{p}_{z,z'}(y).$
\end{definition}

Then, we introduce a Lemma as follows,
\begin{lemma}\label{lemma:wasserstein}
    \cite{edwards2011kantorovich}.
    A function \( h \) is called Lipschitz with respect to \( m \) if \( |h(x) - h(y)| \leq m(x, y) \) for all \( x, y \), and let \( \mathcal{H}(m) \) denote the space of such functions. If \( m \) is a metric, the Wasserstein distance can be expressed as:  
\[
W_c(p_{z,z'}, \widehat{p}_{z,z'}) = \sup_{h \in \mathcal{H}(m)} \mathbb{E}_{X \sim p_{z,z'}}[h(X)] - \mathbb{E}_{X \sim \widehat{p}_{z,z'}}[h(X)].
\]  
Now, let \( m(x, y) = \mathbb{I}_{x \neq y} \). In this case, the Total Variation (TV) distance becomes:  
\[
TV(p_{z,z'}, \widehat{p}_{z,z'}) = \sup_{h : \mathcal{X} \to [0, 1]} \mathbb{E}_{X \sim p_{z,z'}}[h(X)] - \mathbb{E}_{X \sim \widehat{p}_{z,z'}}[h(X)].
\]
\end{lemma}

Finally, we prove Theorem \ref{theorem:fairness_gap} as follows,
\begin{proof}
For any pairwise group labels $z, z'$,
\begin{equation*}   
\begin{aligned}
    g_{z, z'}(\theta) &= g_{z, z'}(\theta) - \widehat{g}_{z, z'}(\theta) + \widehat{g}_{z, z'}(\theta) \leq |g_{z, z'}(\theta) - \widehat{g}_{z, z'}(\theta)| + \widehat{g}_{z, z'}(\theta).
\end{aligned}
\end{equation*}
By Lemma \ref{lemma:wasserstein}, we have the following result.
\begin{equation*}   
\begin{aligned}
    |g_{z, z'}(\theta)-\hat{g}_{z, z'}(\theta)| = \big|\mathbb{E}[h(\theta)|Z=z, Z'=z']- \mathbb{E}[h(\theta)|\widehat{Z}=z, \widehat{Z}'=z']\big| \leq TV(p_{z, z'}, \hat{p}_{z, z'}).
\end{aligned}
\end{equation*}

Given the assumption that \( \theta \) satisfies the fairness constraints with respect to the noisy groups, \( \hat{g}_{z, z'}(\theta) \leq 0 \). Therefore, we derive the desired result:  
\[
g_{z, z'}(\theta) \leq TV(p_{z, z'}, \hat{p}_{z, z'}) \leq \gamma_{z, z'}.
\]
    
\end{proof}



\subsection{Proof of Lemma \ref{lemma:estimate_ratio}}\label{appendix:proof_lemma}


\textit{Proof.} The Total Variation (TV) distance between the probability measures \( p_{z, z'} \) and \( \hat{p}_{z, z'} \) is defined as:
\[
\begin{aligned}
  &TV(p_{z, z'}, \hat{p}_{z, z'}) = \sup \{ |p_{z, z'}(A) - \hat{p}_{z, z'}(A)| : A \text{ is a measurable event} \}. 
\end{aligned}
\]

Let \( A \) be any measurable event under both \( p_{z, z'} \) and \( \hat{p}_{z, z'} \). By the definition of \( p_{z, z'} \), we have \( p_{z, z'}(A) = Pr[A \mid (Z, Z') = (z, z')] \). In the context of the pairwise problem between \( p_{z, z'} \) and \( \hat{p}_{z, z'} \), it follows that: 
\begin{align*}
& |p_{z, z'}(A) - \widehat{p}_{z, z'}(A)| \\
&= |Pr[A \mid (Z, Z') = (z,z')] - Pr[A \mid (\widehat{Z}, \widehat{Z}')=(z,z')]| \\
&= |Pr[A \mid (Z, Z') = (z,z'), (\widehat{Z}, \widehat{Z}')=(z,z')]Pr[(\widehat{Z}, \widehat{Z}')=(z,z') \mid (Z, Z') = (z,z')] \\
&\quad + Pr[A \mid (Z, Z') = (z,z'), (\widehat{Z}, \widehat{Z}')\neq(z,z')]Pr[((\widehat{Z}, \widehat{Z}') \neq (z, z') \mid (Z, Z') = (z,z')] \\
&\quad - Pr[A \mid (\widehat{Z}, \widehat{Z}')=(z,z'), (Z, Z') = (z,z')]Pr[(Z, Z') = (z,z') \mid (\widehat{Z}, \widehat{Z}')=(z,z')] \\
&\quad - Pr[A \mid (\widehat{Z}, \widehat{Z}')=(z,z'), (Z, Z') \neq (z,z')]Pr[(Z, Z') \neq (z,z') \mid (\widehat{Z}, \widehat{Z}')=(z,z')]| \\
&= |Pr[A \mid (Z, Z') = (z,z'), (\widehat{Z}, \widehat{Z}')=(z,z')] \cdot  \\
&\ \ \ \ \ \ \ \ \ \ \ \ \ \  \big( Pr[(\widehat{Z}, \widehat{Z}')=(z,z') \mid (Z, Z') = (z,z')] -  Pr[(Z, Z') = (z,z') \mid (\widehat{Z}, \widehat{Z}')=(z,z')]\big) \\
&\quad - Pr[(\widehat{Z}, \widehat{Z}') \neq (Z, Z')  \mid (Z, Z') = (z, z')] \cdot \\
&\ \ \ \ \ \ \ \ \ \ \ \ \ \ \big( Pr[A \mid (Z, Z') = (z,z'), (\widehat{Z}, \widehat{Z}')\neq(z,z')] - Pr[A \mid (\widehat{Z}, \widehat{Z}')=(z,z'), (Z, Z') \neq (z,z')] \big)| 
\\
&= |0 - Pr[(\widehat{Z}, \widehat{Z}') \neq (Z, Z')  \mid (Z, Z') = (z, z')] \cdot \\
&\ \ \ \ \ \ \ \ \ \ \ \ \ \ \big( Pr[A \mid (Z, Z') = (z,z'), (\widehat{Z}, \widehat{Z}')\neq(z,z')] - Pr[A \mid (\widehat{Z}, \widehat{Z}')=(z,z'), (Z, Z') \neq (z,z')] \big)| \\
&\leq Pr[(\widehat{Z}, \widehat{Z}') \neq (Z, Z')  \mid (Z, Z') = (z, z')]=Pr[(Z, Z') \neq (\widehat{Z}, \widehat{Z}') \mid (\widehat{Z}, \widehat{Z}') = (z, z')].
\end{align*}
The second equality follows from the law of total probability. The third and the fourth equalities follow from the assumption that $Pr[(Z, Z') = (z, z')]=Pr[(\widehat{Z}, \widehat{Z}') = (z, z')]$, which implies that $Pr[(\widehat{Z}, \widehat{Z}') = (Z, Z')  \mid (Z, Z') = (z, z')]= Pr[(Z, Z')= (\widehat{Z}, \widehat{Z}') \mid (\widehat{Z}, \widehat{Z}') = (z, z')]$ since 
\begin{equation*}
\begin{aligned}
    Pr[(\widehat{Z}, \widehat{Z}') = (Z, Z')  \mid (Z, Z') = (z, z')] &=\frac{Pr[(\widehat{Z}, \widehat{Z}') = (Z, Z'); (Z, Z') = (z, z')]}{Pr[(Z, Z') = (z, z')]}\\
    &=\frac{Pr[(\widehat{Z}, \widehat{Z}') = (Z, Z'); (\widehat{Z}, \widehat{Z}') = (z, z')]}{Pr[(\widehat{Z}, \widehat{Z}') = (z, z')]}\\
    &=Pr[(\widehat{Z}, \widehat{Z}') = (Z, Z')  \mid (\widehat{Z}, \widehat{Z}') = (z, z')]
\end{aligned}
\end{equation*}
This further implies that $Pr[(\widehat{Z}, \widehat{Z}') \neq (Z, Z')  \mid (Z, Z') = (z, z')]=Pr[(\widehat{Z}, \widehat{Z}') \neq (Z, Z')  \mid (\widehat{Z}, \widehat{Z}') = (z, z')]$, which is the reason why the last equation holds.

\section{Sampling Method}\label{append:sampling}
We denote strata of data as $\mathcal{S}^{zY}$, where $z \in \mathcal{Z}$ and $Y \in \mathcal{Y}$. Then, the sampling algorithm is shown in Algorithm \ref{alg:sampler}.  
\begin{algorithm}[t]
\caption{\sampler(Dataset: $\mathcal{S}$, batch\_size: $b$)}
\label{alg:sampler}
\begin{algorithmic}[1]
\FOR{$z\in\mathcal{Z}$ and $Y\in\mathcal{Y}$}
  \STATE Uniformly sample without replacement $B^{zY}$ from $\mathcal{S}^{zY}$ with size $b^{zY}=\lceil b\cdot(|\mathcal{S}^{zY}|/|S|)\rceil$
\ENDFOR
\STATE \textbf{return} $B=\cup_{z,Y}B^{zY}$
\end{algorithmic}
\end{algorithm}

\section{Datasets Details}\label{appendix:dataset}
\begin{table}[h]
\centering
\begin{tabular}{ccccc}
\hline
\textbf{Type}            & \textbf{Name} & \textbf{\# Instances} & \textbf{Group ratio} & \textbf{Class ratio} \\ \hline
\multirow{3}{*}{Tabular} & Adult         & 48,842                & 0.48:1               & 3.03:1               \\
                         & Bank          & 41,188                & 0.05:1               & 7.55:1               \\
                         & Default       & 30,000                & 1.52:1               & 3.52:1               \\ \hline
\multirow{4}{*}{Image}   & FF++          & 76,139                & 1.27:1               & 1:4.89               \\
                         & DFDC          & 22,857                & 1.05:1               & 1.80:1               \\
                         & DFD           & 9,386                 & 1.22:1               & 1:2.04               \\
                         & Celeb-DF      & 28,458                & 8.79:1               & 1:3.98               \\ \hline
\end{tabular}
\caption{Dataset Statistics. The group ratio is given by the protective attribute 
\(Z = a\) vs.\ \(Z = b\). The class ratio is given by negative vs.\ positive class.}
\label{tab:dataset-stats}
\end{table}
For tabular datasets, we do socioeconomic analysis on three datasets that have been commonly used in the fair machine learning literature \cite{donini2018empirical}. In \texttt{Adult} dataset, the sensitive attribute is the gender of the individual, i.e.\ female ($Z = a$) or male ($Z = b$). In \texttt{Bank} dataset , the sensitive attribute is the age of the individual: $Z = a$ when the age is less than 25 or over 60 and $Z = b$ otherwise. In \texttt{Default} dataset \cite{yeh2009comparisons} , the sensitive attribute is the gender of the individual, i.e. female ($Z = a$) or male ($Z = b$). 

For image datasets, we do deepfake detection on the most widely used benchmark FaceForensics++(FF++) \cite{rossler2019faceforensics++}. DFDC~\cite{deepfakedetection2021}, DFD~\cite{googledeepfakes2019}, and Celeb-DF~\cite{li2020celebdf}. We only used the test set of the later three deepfake dataset. Since the original datasets do not have the demographic information of each video or image, we follow \citet{ju2024improving} for data processing and data annotation. The sensitive attribute is the gender of the individual, i.e. female ($Z = a$) or male ($Z = b$). The summary statistics of the datasets are given in Table \ref{tab:dataset-stats}.

\section{Fairness Metrics}\label{appendix:metric}
For fairness, we measure AUC fairness violation (`Violation', lower is better), defined as the maximum absolute difference between any group-level AUC score the overall AUC score: 
\begin{equation*}
    \max_{z,z'\in \mathcal{Z}}|g_{z,z'}(\theta)|.
\end{equation*}
Additionally, following \citet{yang2023minimax}, we use the Min/Max fairness metric (higher is better), defined as the ratio of the minimum to the maximum group-level AUC score: 
\begin{equation*}
    \min_{z,z'\in\mathcal{Z}}AUC_{z,z'}(\theta)/\max_{z,z'\in\mathcal{Z}}AUC_{z,z'}(\theta).
\end{equation*}

\section{Baselines}\label{appendix:baselines}

\begin{itemize}
    \item The AUCMax algorithm maximizes AUC across the entire dataset without distinguishing between groups. It updates the model parameters using mini-batch SGD.
    \item We select the method by \citet{vogel2021learning} as a representative since they considered the same datasets. Their approach, which we refer to as InterFairAUC, ensures fair AUC scores by regularizing the difference between inter-group AUCs.
    \item MinimaxFairAUC introduced by \citet{yang2023minimax} is a minimax fairness
framework that simultaneously addresses intra-group and
inter-group AUC disparities using a Rawlsian approach, sup-
ported by an efficient optimization algorithm with proven
convergence guarantees.
    \item DAG-FDD~\cite{ju2024improving}, a demographic-aware Fair Deepfake Detection (DAW-FDD) method leverages demographic information and employs an existing fairness risk measure~\cite{williamson2019fairness}. At a high level, DAW-FDD aims to ensure that the losses achieved by different user-specified groups of interest (\eg, different races or genders) are similar to each other (so that the AI face detector is not more accurate on one group vs another) and, moreover, that the losses across all groups are low.  Specifically, DAW-FDD uses a CVaR~\cite{levy2020large, rockafellar2000optimization} loss function across groups (to address imbalance in demographic groups) and, per group, DAW-FDD uses another CVaR loss function (to address imbalance in real vs AI-generated training examples).
    \item DAW-FDD~\cite{ju2024improving}, a demographic-agnostic Fair Deepfake Detection (DAG-FDD) method, which is based on the distributionally robust optimization (DRO)~\cite{hashimoto2018fairness, duchi2021learning}. To use DAG-FDD, the user does not have to specify which attributes to treat as sensitive such as race and gender, only need to specify a probability threshold for a minority group without explicitly identifying all possible groups.
    \item PG-FDD~\cite{lin2024preserving} (Preserving Generalization Fair Deepfake Detection) employs disentanglement learning to extract demographic and domain-agnostic forgery features, promoting fair learning across a flattened loss landscape. Its framework combines disentanglement learning, fairness learning, and optimization modules. The disentanglement module introduces a loss to expose demographic and domain-agnostic features that enhance fairness generalization. The fairness learning module combines these features to promote fair learning, guided by generalization principles. The optimization module flattens the loss landscape, helping the model escape suboptimal solutions and strengthen fairness generalization.
\end{itemize}

\section{More Implementation Details and results}\label{appendix:implementation}
\subsection{Additional experimental setup details}\label{appendix:additional_implementation}
In Algorithm \ref{alg:proj-gda}, we explored the following hyperparameters:
\begin{itemize}
    \item Tabular data:
    \begin{enumerate}
        \item $\eta_{\theta} \in \{0.001, 0.01, 0.1\}$
        \item $\eta_{\lambda} \in \{0.001, 0.25, 0.5\}$
        \item $\eta_{p} \in \{0.001, 0.01, 0.1\}$
        \item $\nu$ in Eq. (\ref{eq:sam_optimization}) is selected from $\{0.0005, 0.001, 0.005\}$
    \end{enumerate}
    \item Image data:
    \begin{enumerate}
        \item $\eta_{\theta} \in \{0.0001, 0.0005, 0.001\}$
        \item $\eta_{\lambda} \in \{0.0001, 0.0005, 0.005\}$
        \item $\eta_{p} \in \{0.0001, 0.0005, 0.001\}$
        \item $\nu$ in Eq. (\ref{eq:sam_optimization}) is selected from $\{0.7, 0.5, 0.3\}$
    \end{enumerate}
\end{itemize}

\subsection{Additional experimental results}\label{appendix:more_non_robust_experiments}
\begin{table*}[t]
    \centering
    \scalebox{0.6}{
    \begin{tabular}{c|c|ccc|ccc|ccc}
    \hline
    \multirow{2}{*}{\makecell{Noise \\ Level}} 
    & \multirow{2}{*}{Method} 
    & \multicolumn{3}{c|}{Adult} 
    & \multicolumn{3}{c|}{Bank} 
    & \multicolumn{3}{c}{Default} \\ \cline{3-11} 
     &  & \multicolumn{1}{c|}{AUC$\uparrow$} 
         & Violation$\downarrow$ 
         & Min/Max$\uparrow$ 
       & \multicolumn{1}{c|}{AUC$\uparrow$} 
         & Violation$\downarrow$ 
         & Min/Max$\uparrow$ 
       & \multicolumn{1}{c|}{AUC$\uparrow$} 
         & Violation$\downarrow$ 
         & Min/Max$\uparrow$ \\ 
    \hline
    
\multirow{2}{*}{0.1}
&  Ours(Non-robust) 
& \multicolumn{1}{c|}{ 0.9064$\pm$0.0008}
&  0.0370$\pm$0.0006
&  0.9535$\pm$0.0006
& \multicolumn{1}{c|}{ 0.9059$\pm$0.0001}
&  0.0920$\pm$0.0029
&  0.9102$\pm$0.0025
& \multicolumn{1}{c|}{ 0.7538$\pm$0.0024}
&  0.0212$\pm$0.0016
&  0.9650$\pm$0.0019 \\

&  Ours (Robust) 
& \multicolumn{1}{c|}{ 0.9060$\pm$0.0002}
&  \textbf{0.0332$\pm$0.0007}
&  \textbf{0.9536$\pm$0.0005}
& \multicolumn{1}{c|}{ 0.9039$\pm$0.0007}
&  \textbf{0.0876$\pm$0.0033}
&  \textbf{0.9143$\pm$0.0029}
& \multicolumn{1}{c|}{ 0.7645$\pm$0.0017}
&  \textbf{0.0187$\pm$0.0012}
&  \textbf{0.9691$\pm$0.0019} \\

\hline
\multirow{2}{*}{0.2}

&  Ours(Non-robust) 
& \multicolumn{1}{c|}{ 0.9108$\pm$0.0009}
&  0.0388$\pm$0.0006
&  0.9532$\pm$0.0004
& \multicolumn{1}{c|}{ 0.9107$\pm$0.0008}
&  \textbf{0.0868$\pm$0.0029}
&  \textbf{0.9111$\pm$0.0024}
& \multicolumn{1}{c|}{ 0.7511$\pm$0.0016}
&  0.0230$\pm$0.0018
&  0.9626$\pm$0.0019 \\

&  Ours (Robust) 
& \multicolumn{1}{c|}{ 0.9039$\pm$0.0004}
&  \textbf{0.0328$\pm$0.0008}
&  \textbf{0.9539$\pm$0.0011}
& \multicolumn{1}{c|}{ 0.9163$\pm$0.0005}
&  0.1053$\pm$0.0026
&  0.8988$\pm$0.0024
& \multicolumn{1}{c|}{ 0.7624$\pm$0.0016}
&  \textbf{0.0206$\pm$0.0014}
&  \textbf{0.9660$\pm$0.0022} \\

\hline
\multirow{2}{*}{0.3}

&  Ours(Non-robust) 
& \multicolumn{1}{c|}{ 0.8971$\pm$0.0072}
&  0.0370$\pm$0.0037
&  \textbf{0.9569$\pm$0.0007}
& \multicolumn{1}{c|}{ 0.9060$\pm$0.0012}
&  0.0973$\pm$0.0026
&  0.9053$\pm$0.0025
& \multicolumn{1}{c|}{ 0.7540$\pm$0.0025}
&  0.0213$\pm$0.0015
&  0.9647$\pm$0.0019 \\

&  Ours (Robust) 
& \multicolumn{1}{c|}{ 0.9059$\pm$0.0005}
&  \textbf{0.0356$\pm$0.0006}
&  0.9513$\pm$0.0007
& \multicolumn{1}{c|}{ 0.9093$\pm$0.0004}
&  \textbf{0.0949$\pm$0.0026}
&  \textbf{0.9080$\pm$0.0024}
& \multicolumn{1}{c|}{ 0.7687$\pm$0.0011}
&  \textbf{0.0129$\pm$0.0011}
&  \textbf{0.9785$\pm$0.0014} \\

\hline
\multirow{2}{*}{0.4}

&  Ours(Non-robust) 
& \multicolumn{1}{c|}{ 0.9075$\pm$0.0012}
&  0.0377$\pm$0.0006
&  0.9522$\pm$0.0001
& \multicolumn{1}{c|}{ 0.9109$\pm$0.0011}
&  0.0943$\pm$0.0010
&  0.9085$\pm$0.0008
& \multicolumn{1}{c|}{ 0.7429$\pm$0.0018}
&  0.0263$\pm$0.0007
&  \textbf{0.9594$\pm$0.0015} \\

&  Ours (Robust) 
& \multicolumn{1}{c|}{ 0.9008$\pm$0.0002}
&  \textbf{0.0341$\pm$0.0005}
&  \textbf{0.9530$\pm$0.0004}
& \multicolumn{1}{c|}{ 0.9054$\pm$0.0004}
&  \textbf{0.0937$\pm$0.0034}
&  \textbf{0.9086$\pm$0.0028}
& \multicolumn{1}{c|}{ 0.7552$\pm$0.0024}
&  \textbf{0.0246$\pm$0.0036}
&  0.9580$\pm$0.0066 \\

\hline
\multirow{2}{*}{0.5}

&  Ours(Non-robust) 
& \multicolumn{1}{c|}{ 0.8989$\pm$0.0011}
&  0.0372$\pm$0.0007
&  0.9553$\pm$0.0002
& \multicolumn{1}{c|}{ 0.9108$\pm$0.0012}
&  0.0895$\pm$0.0029
&  0.9126$\pm$0.0024
& \multicolumn{1}{c|}{ 0.7505$\pm$0.0014}
&  0.0261$\pm$0.0001
&  0.9564$\pm$0.0001 \\

&  Ours (Robust) 
& \multicolumn{1}{c|}{ 0.8984$\pm$0.0006}
&  \textbf{0.0316$\pm$0.0002}
&  \textbf{0.9554$\pm$0.0004}
& \multicolumn{1}{c|}{ 0.9044$\pm$0.0004}
&  \textbf{0.0876$\pm$0.0020}
&  \textbf{0.9145$\pm$0.0015}
& \multicolumn{1}{c|}{ 0.7447$\pm$0.0047}
&  \textbf{0.0243$\pm$0.0014}
&  \textbf{0.9571$\pm$0.0010} \\

\hline
\end{tabular}
}
\caption{ 
\small \textit{Performance comparison across different noise levels (0.1--0.5) between robust and non-robust method of ours. The numbers are reported as `Mean $\pm$ Standard Deviation.’ \textuparrow~means higher is better and \textdownarrow~means lower is better.}}
\label{tab:tabular_more}
\end{table*}

\begin{table*}[t]
    \centering
    \scalebox{0.64}{
\begin{tabular}{c|c|ccc|ccc|ccc|ccc}
\hline
                                  &                                          & \multicolumn{3}{c|}{FF++}                                                                                                                        & \multicolumn{3}{c|}{DFDC}                                                                                                                        & \multicolumn{3}{c|}{DFD}                                                                                                                         & \multicolumn{3}{c}{Celeb-DF}                                                                                                            \\ \cline{3-14} 
\multirow{-2}{*}{Backbone}        & \multirow{-2}{*}{Method}                 & \multicolumn{1}{c|}{AUC\textuparrow}                                     & Violation\textdownarrow                               & Min/Max\textuparrow                                 & \multicolumn{1}{c|}{AUC\textuparrow}                                     & Violation\textdownarrow                               & Min/Max\textuparrow                                 & \multicolumn{1}{c|}{AUC\textuparrow}                                     & Violation\textdownarrow                               & Min/Max\textuparrow                                 & \multicolumn{1}{c|}{AUC\textuparrow}                            & Violation\textdownarrow                               & Min/Max\textuparrow                                 \\ \hline
                                  & Ours(Non-robust) & \multicolumn{1}{c|}{0.9546}          & 0.0103          & 0.9842          & \multicolumn{1}{c|}{0.6014}          & 0.0191          & 0.9573          & \multicolumn{1}{c|}{0.7826}          & 0.0098          & 0.9863          & \multicolumn{1}{c|}{0.7105} & 0.0994          & 0.8807          \\
\multirow{-2}{*}{Xception}        & Ours    & \multicolumn{1}{c|}{0.9644}          & \textbf{0.0090} & \textbf{0.9857} & \multicolumn{1}{c|}{0.6086}          & \textbf{0.0048} & \textbf{0.9930} & \multicolumn{1}{c|}{0.7847}          & \textbf{0.0069} & \textbf{0.9881} & \multicolumn{1}{c|}{0.7108} & \textbf{0.0729} & \textbf{0.9117} \\ \hline
                                 
                                  & Ours(Non-robust) & \multicolumn{1}{c|}{0.9729}          & 0.0100          & 0.9842          & \multicolumn{1}{c|}{0.6090}          & 0.0168          & 0.9636          & \multicolumn{1}{c|}{0.8058}          & 0.0289          & 0.9611          & \multicolumn{1}{c|}{0.7330} & 0.1046          & 0.8715          \\
\multirow{-2}{*}{EfficientNet-B4} & Ours     & \multicolumn{1}{c|}{\textbf{0.9766}} & \textbf{0.0061} & \textbf{0.9907} & \multicolumn{1}{c|}{\textbf{0.6172}} & \textbf{0.0136} & \textbf{0.9771} & \multicolumn{1}{c|}{\textbf{0.8184}} & \textbf{0.0135} & \textbf{0.9760} & \multicolumn{1}{c|}{0.7351} & \textbf{0.0928} & \textbf{0.8876} \\ \hline
\end{tabular}
}
\caption{\small \textit{Performance comparison of Ours (Non-robust) and Ours on deepfake detection task. \textuparrow~means higher is better and \textdownarrow~means lower is better.}}
\label{tab:deepfake-more}
\end{table*}

Table \ref{tab:tabular_more} and table \ref{tab:deepfake-more} show the results between robust and non-robust methods of ours. In general, our method with using robust approach shows more robust performance than our method without the robust approach.

\begin{table}[t]
\centering
\begin{tabular}{c|c|ccc}
\hline
\multirow{2}{*}{Backbone} & \multicolumn{1}{c|}{\multirow{2}{*}{Method}} & \multicolumn{3}{c}{FF++ clean} \\ \cline{3-5} 
                          & \multicolumn{1}{c|}{}                        & AUC\textuparrow     & Violation\textdownarrow   & Min/Max\textdownarrow   \\ \hline
\multirow{5}{*}{Xception} & Ori                                          & 0.9384  & 0.0289     & 0.9691   \\
                          & DAG-FDD                                      & 0.9628  & 0.0128     & 0.9808   \\
                          & DAW-FDD                                      & 0.965   & 0.0256     & 0.9751   \\
                          & PG-FDD                                       & \textbf{0.9708}  & 0.0105     & 0.9840    \\
                          & \cellcolor[HTML]{EFEFEF}Ours                                         & \cellcolor[HTML]{EFEFEF}0.9644  & \cellcolor[HTML]{EFEFEF} \textbf{0.0070}      & \cellcolor[HTML]{EFEFEF} \textbf{0.9894}   \\ \hline
\end{tabular}
\caption{\small \textit{Performance comparison on clean version of FF++ using Xception backbone. \textuparrow~means higher is better and \textdownarrow~means lower is better.}}
\label{tab:clean}
\end{table}

\textbf{Label setting.} In Table \ref{tab:tabular}, fairness metrics use underlying group labels. In Table \ref{tab:deepfake}, we use noisy group labels, which are common in datasets like FF++ where demographic attributes are inferred. Evaluating fairness under label noise is practical and follows prior work (\cite{celis2021fair, mehrotra2022fair}). We also test the model on a human-corrected FF++ test set from the \citet{lin2024ai}, which contains relatively clean labels. As shown in Table \ref{tab:clean}, our method still maintains the best performance across all fairness metrics.

\begin{table*}[t]
\centering
\begin{tabular}{c|c|c}
\hline
Backbone                  & Method   & Training Time per Epoch (minutes) \\ \hline
\multirow{5}{*}{Xception} & Original & 8                                 \\
                          & DAG-FDD  & 6                                 \\
                          & DAW-FDD  & 9                                 \\
                          & PG-FDD   & 28                                \\
                          & Ours     & 15                                \\ \hline
\end{tabular}
\caption{\small \textit{Training time comparison on deepfake detection task on FF++ dataset.}}
\label{tab:time}
\end{table*}

\textbf{Computational overhead.} To evaluate the practicality of our approach at scale, we benchmarked training time on the FaceForensics++ dataset, a widely used, large-scale benchmark for deepfake detection. As shown in table \ref{tab:time}, our method introduces moderate overhead compared to some baselines but remains significantly more efficient than others, such as PG-FDD. Specifically, our method requires 15 minutes per epoch, which is faster than PG-FDD (28 min) and reasonably close to other baselines like DAW-FDD and the Original model. This demonstrates that our approach remains computationally feasible and scalable in practice, even for large image datasets and backbone models like Xception.

\begin{table}
    \centering{
    \begin{tabular}{c|c|ccc}
    \hline
    \multirow{2}{*}{\makecell{Noise \\ Level}} 
    & \multirow{2}{*}{Method} 
    & \multicolumn{3}{c}{Default} 
   \\ \cline{3-5} 
     &  & \multicolumn{1}{c|}{AUC$\uparrow$} 
         & Violation$\downarrow$ 
         & Min/Max$\uparrow$ \\ 
    \hline
    
\multirow{4}{*}{0.6}

& AUCMax 
& \multicolumn{1}{c|}{\textbf{0.7771$\pm$0.0005}}
& 0.0672$\pm$0.0004 
& 0.9108$\pm$0.0005
 \\

& InterFairAUC 
& \multicolumn{1}{c|}{0.7548$\pm$0.0022}
& 0.0238$\pm$0.0039
& 0.9586$\pm$0.0062
 \\

& MinimaxFairAUC 
& \multicolumn{1}{c|}{0.7550$\pm$0.0021}
& 0.0249$\pm$0.0054
& 0.9577$\pm$0.0078
\\

& \cellcolor[HTML]{EFEFEF}Ours
& \multicolumn{1}{c|}{\cellcolor[HTML]{EFEFEF}0.7540$\pm$0.0024}
& \cellcolor[HTML]{EFEFEF}\textbf{0.0186$\pm$0.0027}
& \cellcolor[HTML]{EFEFEF}\textbf{0.9676$\pm$0.0045}
\\

\hline
\multirow{4}{*}{0.7}

& AUCMax 
& \multicolumn{1}{c|}{\textbf{0.7788$\pm$0.0007}}
& 0.0696$\pm$0.0010 
& 0.9081$\pm$0.0012
 \\

& InterFairAUC 
& \multicolumn{1}{c|}{0.7532$\pm$0.0020}
& 0.0239$\pm$0.0018
& 0.9594$\pm$0.0052
\\

& MinimaxFairAUC 
& \multicolumn{1}{c|}{0.7545$\pm$0.0021}
& 0.0265$\pm$0.0062
& 0.9557$\pm$0.0079
\\

& \cellcolor[HTML]{EFEFEF}Ours 
& \multicolumn{1}{c|}{\cellcolor[HTML]{EFEFEF}0.7541$\pm$0.0024}
& \cellcolor[HTML]{EFEFEF}\textbf{0.0183$\pm$0.0033}
& \cellcolor[HTML]{EFEFEF}\textbf{0.9687$\pm$0.0052}
\\

\hline
\multirow{4}{*}{0.8}

& AUCMax 
& \multicolumn{1}{c|}{\textbf{0.7760$\pm$0.0006}}
& 0.0687$\pm$0.0014
& 0.9092$\pm$0.0017
\\

& InterFairAUC 
& \multicolumn{1}{c|}{0.7548$\pm$0.0022}
& 0.0251$\pm$0.0043
& 0.9566$\pm$0.0062
\\

& MinimaxFairAUC 
& \multicolumn{1}{c|}{0.7532$\pm$0.0021}
& 0.0246$\pm$0.0039
& 0.9588$\pm$0.0072
\\

& \cellcolor[HTML]{EFEFEF}Ours
& \multicolumn{1}{c|}{\cellcolor[HTML]{EFEFEF}0.7546$\pm$0.0025}
& \cellcolor[HTML]{EFEFEF}\textbf{0.0183$\pm$0.0035}
& \cellcolor[HTML]{EFEFEF}\textbf{0.9688$\pm$0.0057}
\\

\hline
\multirow{4}{*}{0.9}

& AUCMax 
& \multicolumn{1}{c|}{\textbf{0.7795$\pm$0.0006}}
& 0.0675$\pm$0.0007
& 0.9108$\pm$0.0008
\\

& InterFairAUC 
& \multicolumn{1}{c|}{0.7545$\pm$0.0022}
& 0.0252$\pm$0.0045
& 0.9567$\pm$0.0062
\\

& MinimaxFairAUC 
& \multicolumn{1}{c|}{0.7544$\pm$0.0022}
& 0.0267$\pm$0.0061
& 0.9555$\pm$0.0078
\\

& \cellcolor[HTML]{EFEFEF}Ours
& \multicolumn{1}{c|}{\cellcolor[HTML]{EFEFEF}0.7527$\pm$0.0022}
& \cellcolor[HTML]{EFEFEF}\textbf{0.0201$\pm$0.0031}
& \cellcolor[HTML]{EFEFEF}\textbf{0.9661$\pm$0.0049}
\\

\hline

\end{tabular}
}
\caption{ 
\small \textit{Performance comparison across high noise levels (0.6--0.9). The numbers are reported as `Mean $\pm$ Standard Deviation.’ \textuparrow~means higher is better and \textdownarrow~means lower is better.  The best results are shown in \textbf{Bold}.}}
\label{tab:noise}
\end{table}

\textbf{Experiments with extreme settings.}  As Table \ref{tab:noise} shows, even in extremely high noise levels (60\%, 70\%, 80\%, and 90\%), our method consistently achieves the lowest AUC fairness violation and the highest Min/Max AUC score. These results strongly align with our claims in the paper, demonstrating that our approach maintains robust fairness guarantees and balanced group performance in highly noisy settings.

\begin{table}[t]
\centering
\begin{tabular}{c|c|ccc}
\hline
\multirow{2}{*}{Backbone} & \multicolumn{1}{c|}{\multirow{2}{*}{Method}} & \multicolumn{3}{c}{FF++} \\ \cline{3-5} 
                          & \multicolumn{1}{c|}{}                        & AUC\textuparrow     & Violation\textdownarrow   & Min/Max\textdownarrow   \\ \hline
\multirow{5}{*}{Xception} & Ori                                          & 0.9445  & 0.0335     & 0.9633   \\
                          & DAG-FDD                                      & 0.9647  & 0.0226     & 0.9798   \\
                          & DAW-FDD                                      & 0.9670   & 0.0215     & 0.9753   \\
                          & PG-FDD                                       & \textbf{0.9751}  & 0.0172     & 0.9837    \\
                          & \cellcolor[HTML]{EFEFEF}Ours                                         & \cellcolor[HTML]{EFEFEF}0.9625  & \cellcolor[HTML]{EFEFEF} \textbf{0.0161}      & \cellcolor[HTML]{EFEFEF} \textbf{0.9850}   \\ \hline
\end{tabular}
\caption{\small \textit{Performance comparison on the FF++ dataset with race as the protected attribute. \textuparrow~means higher is better and \textdownarrow~means lower is better.}}
\label{tab:race}
\end{table}

\textbf{Results on more groups.} We add one more experiment about evaluating our method on a dataset with more than two protected groups. Specifically, in Table \ref{tab:race}, we present results on the FaceForensics++ (FF++) dataset, where we consider race as the protected attribute, comprising four groups: White, Black, Asian, and Other. This multi-group setting is more challenging than the binary group setting commonly seen in fairness literature. Nonetheless, our method achieves the lowest fairness violation (0.0161) and the highest Min/Max AUC score (0.9850) among all baselines, demonstrating its effectiveness in ensuring AUC fairness across multiple protected groups. These results confirm that our approach generalizes well to settings involving complex, non-binary group structures, such as race.

\begin{table}[t]
\centering
\begin{tabular}{c|c|ccc}
\hline
\multirow{2}{*}{Backbone} & \multicolumn{1}{c|}{\multirow{2}{*}{Method}} & \multicolumn{3}{c}{FF++} \\ \cline{3-5} 
                          & \multicolumn{1}{c|}{}                        & AUC\textuparrow     & Violation\textdownarrow   & Min/Max\textdownarrow   \\ \hline
\multirow{5}{*}{EfficientNet-B4} & Ori                                          & 0.9332  & 0.0209     & 0.9684   \\
                          & DAG-FDD                                      & 0.9563  & 0.0100     & 0.9869   \\
                          & DAW-FDD                                      & 0.9694   & 0.0169     & 0.9764   \\
                          & PG-FDD                                       & 0.9721  & 0.0144     & 0.9784    \\
                          & \cellcolor[HTML]{EFEFEF}Ours(CLIP-labeled)                                       & \cellcolor[HTML]{EFEFEF}0.9725  & \cellcolor[HTML]{EFEFEF}0.0089     & \cellcolor[HTML]{EFEFEF}0.9858    \\
                          & \cellcolor[HTML]{EFEFEF}Ours(robust)                                         & \cellcolor[HTML]{EFEFEF}\textbf{0.9766}  & \cellcolor[HTML]{EFEFEF} \textbf{0.0061}      & \cellcolor[HTML]{EFEFEF} \textbf{0.9907}   \\ \hline
\end{tabular}
\caption{\small \textit{Performance comparison on FF++ using EfficientNet-B4 backbone. \textuparrow~means higher is better and \textdownarrow~means lower is better.}}
\label{tab:clip_baseline}
\end{table}

\textbf{CLIP for label prediction.} We added experiments with one more baseline to address your concern. Specifically, as shown in Table \ref{tab:clip_baseline}, we include a baseline called Ours (CLIP-labeled), where we directly use CLIP to predict the protected group labels and then train the model based on those labels. In contrast, our full method, Ours (robust), uses CLIP only to estimate the group label noise level, not for prediction or relabeling. It is clear that Ours (robust) outperforms Ours (CLIP-labeled) in all metrics, achieving higher AUC (0.9766 vs. 0.9725), lower violation (0.0061 vs. 0.0089), and better Min/Max fairness (0.9907 vs. 0.9858). This indicates that directly using CLIP-predicted labels to mitigate the impact of noise during training does not yield optimal performance. More importantly, this baseline approach does not guarantee fairness under the noisy label setting.

\begin{figure*}[t]
  \centering
  \includegraphics[width=1.0\linewidth]{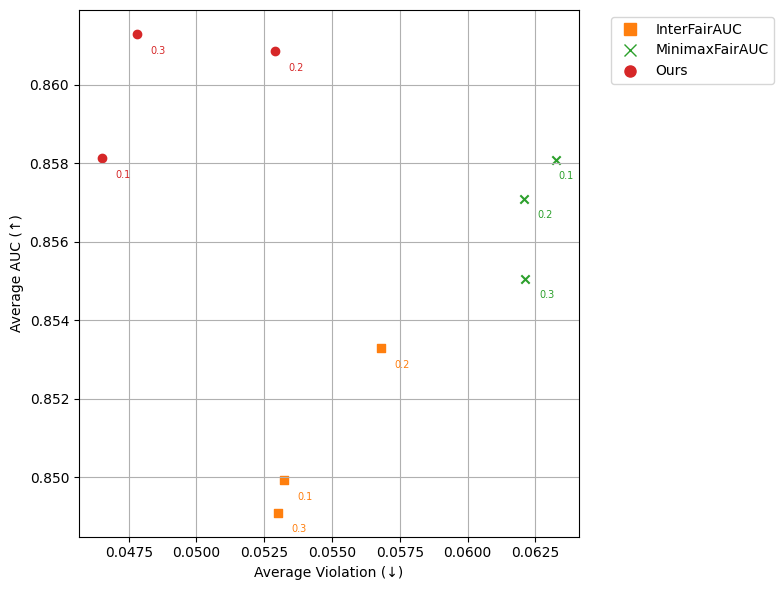}
  \caption{\it \small Efficiency frontier showing the trade-off between Average AUC and Average Fairness Violation across three tabular datasets (Adult, Bank, Default) at varying noise levels (0.1–0.3). }
  \vspace{-2mm}
  \label{fig:auc_vio1}
\end{figure*}

\begin{figure*}[t]
  \centering
  \includegraphics[width=1.0\linewidth]{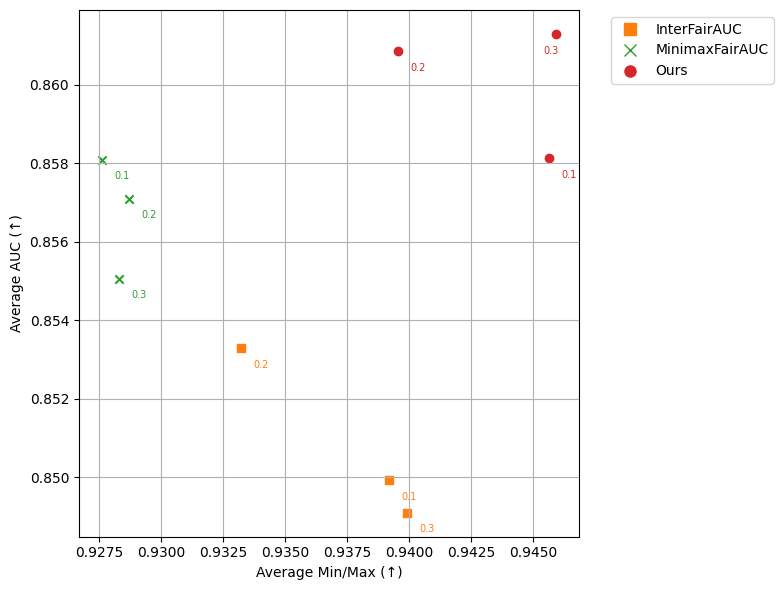}
  \caption{\it \small Efficiency frontier showing the trade-off between Average AUC and Average Min/Max AUC across three tabular datasets (Adult, Bank, Default) at varying noise levels (0.1–0.3). }
  \vspace{-2mm}
  \label{fig:auc_minimax1}
\end{figure*}

\begin{figure*}[t]
  \centering
  \includegraphics[width=1.0\linewidth]{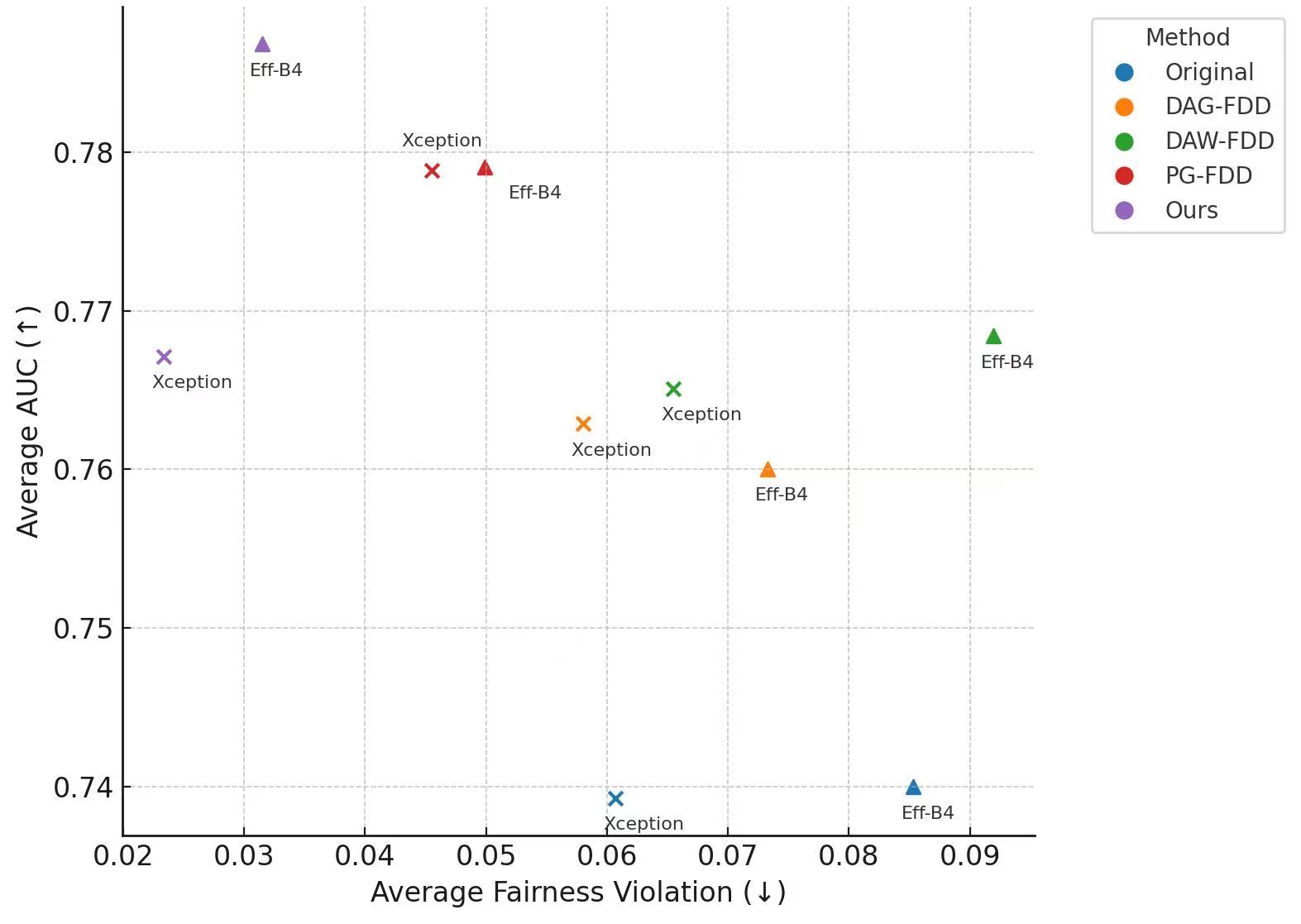}
  \caption{\it \small Efficiency frontier showing the trade-off between detection performance (AUC) and fairness violation across four benchmark datasets (FF++, DFDC, DFD, Celeb-DF). }
  \vspace{-2mm}
  \label{fig:auc_vio2}
\end{figure*}

\begin{figure*}[t]
  \centering
  \includegraphics[width=1.0\linewidth]{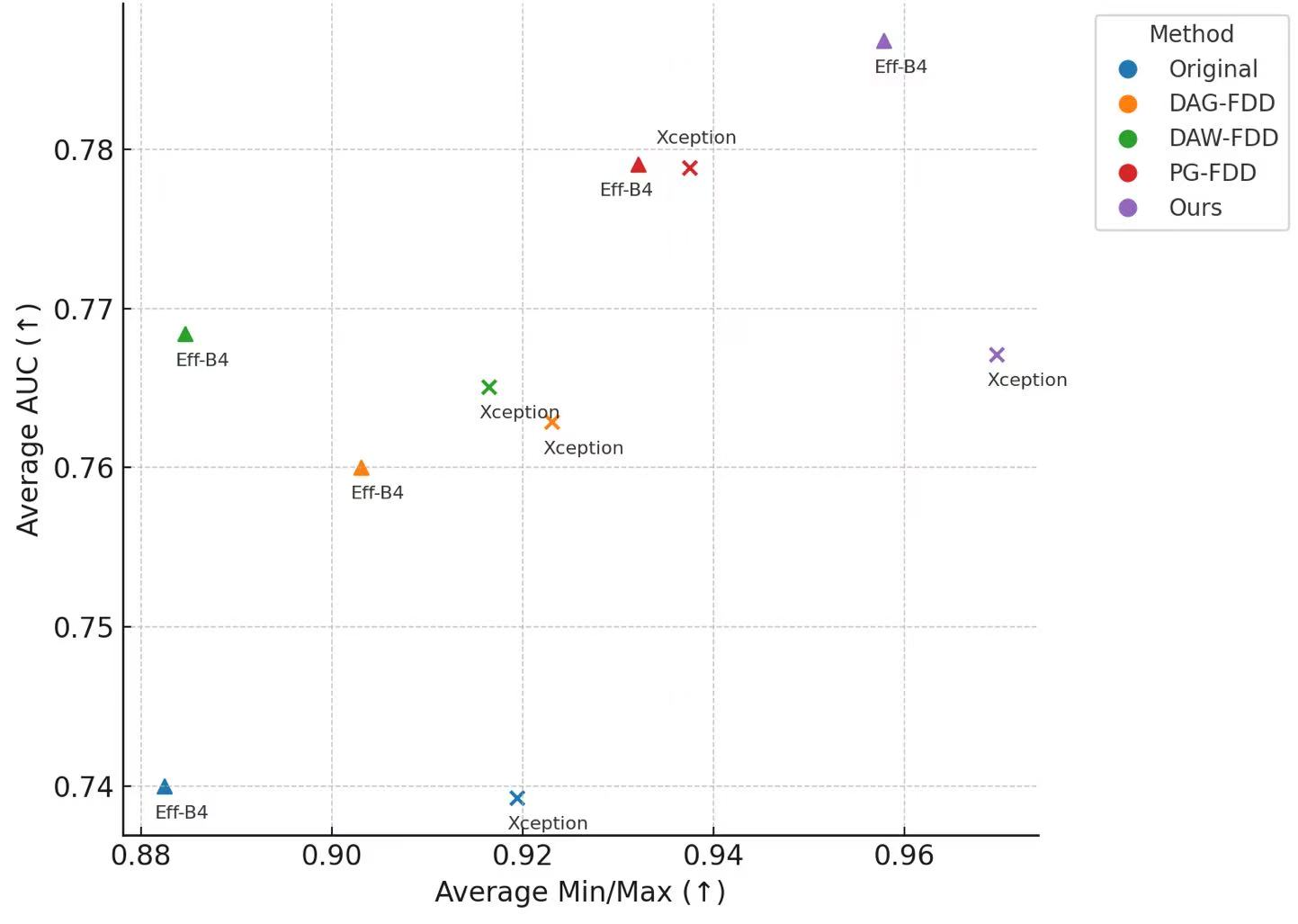}
  \caption{\it \small Efficiency frontier showing the trade-off between detection performance (AUC) and Min/Max AUC across four benchmark datasets (FF++, DFDC, DFD, Celeb-DF). }
  \vspace{-2mm}
  \label{fig:auc_minimax2}
\end{figure*}

\textbf{Additional Visualizations.} To better communicate the fairness–performance tradeoff across all methods in Table \ref{tab:tabular}, we include two efficiency frontier plots in Fig \ref{fig:auc_vio1} and Fig \ref{fig:auc_minimax1}. We used the results in Table \ref{tab:tabular}, which include performance on three tabular datasets: Adult, Bank, and Default. For each fairness-enhanced method, we compute the average AUC, average fairness violation, and average Min/Max AUC ratio across the three datasets for each noise level (0.1–0.3). This gives three points per method, each representing the average values across datasets at a given noise level. Fig \ref{fig:auc_vio1} visualizes Average AUC vs. Average Fairness Violation, demonstrating the tradeoff of performance and fairness. Our method ("Ours") consistently occupies the top-left region, achieving lower violation than all baselines while maintaining competitive or superior AUC. Fig \ref{fig:auc_minimax1} shows Average AUC vs. Average Min/Max AUC Ratio, highlighting group-wise fairness consistency. Our method ranks in the top-right region, attains the best Min/Max ratio with strong AUC, reflecting better fairness stability across groups.

To complement the tabular results in Table \ref{tab:deepfake}, we include two efficiency frontier plots based on the average values across all four benchmark datasets (FF++, DFDC, DFD, and Celeb-DF). Fig \ref{fig:auc_vio2} reveals the trade-off between detection performance and fairness violation. Our method ("Ours") achieves the lowest average violation while maintaining a higher AUC than all other methods on both Xception and EfficientNet-B4 backbones. These results position our method at the top-left corner, indicating Pareto efficiency and demonstrating strong performance–fairness trade-offs. Fig \ref{fig:auc_minimax2} captures performance versus group-wise fairness consistency. Our method again ranks at the top-right region, with both the highest min/max ratio and competitive or superior AUC. This confirms that our method not only performs well but also offers greater fairness stability across demographic groups. We note that while the table provides fine-grained per-dataset results, these plots offer a complementary view by highlighting global efficiency across multiple objectives. The strong position of our method in both plots further supports its overall effectiveness.

\end{document}